\documentclass[10pt,journal,compsoc]{IEEEtran}
\usepackage{amsmath,amsfonts}
\usepackage{algorithmic}
\usepackage{algorithm}
\usepackage{array}
\usepackage[caption=false,font=normalsize,labelfont=sf,textfont=sf]{subfig}
\usepackage[justification=justified,singlelinecheck=false]{caption}
\usepackage{textcomp}
\usepackage{xcolor}
\usepackage{stfloats}
\usepackage{url}
\usepackage{verbatim}
\usepackage{graphicx}
\usepackage{cite}
\usepackage{newfloat}
\usepackage{listings}
\usepackage{svg}
\usepackage{tikz}
\usepackage{amsmath}
\usepackage{multicol}
\usepackage{multirow}
\usepackage{makecell}
\usepackage{pifont}
\usepackage{adjustbox}
\usepackage{color}
\usepackage{tikz}
\usepackage{overpic}
\usepackage{rotating}
\usepackage{graphicx}
\usepackage{float}
\usepackage{subfig}
\usepackage{ragged2e}
\usepackage{diagbox}
\usepackage{array}
\usepackage{booktabs}
\usepackage{hyperref}
\hypersetup{colorlinks=true,linkcolor=blue,citecolor=blue,urlcolor=blue}

\definecolor{citecolor}{HTML}{2fe062}

\newcommand{\cmark}{\ding{52}}%

\usepackage{cleveref}

\begin{document}

\title{Consolidating Diffusion-Generated Video Detection with Unified Multimodal \\ Forgery Learning}

\author{Xiaohong Liu,~\IEEEmembership{Member,~IEEE},
    Xiufeng Song,
    Huayu Zheng,
    Lei Bai,~\IEEEmembership{Member,~IEEE},\\
    Xiaoming Liu,~\IEEEmembership{Fellow,~IEEE},
    Guangtao Zhai,~\IEEEmembership{Fellow,~IEEE}
\thanks{Xiaohong Liu, Xiufeng Song, Huayu Zheng are with the School of Computer Science, Shanghai Jiao Tong University, Shanghai 200240, China (email: \{xiaohongliu, akikaze, 2539629225\}@sjtu.edu.cn).}
\thanks{Guangtao Zhai is with the School of Information Science and Electronic Engineering, Shanghai Jiao Tong University, Shanghai 200240, China (email: zhaiguangtao@sjtu.edu.cn).}
\thanks{Lei Bai is with the Shanghai Artificial Intelligence Laboratory, Shanghai 200232, China (email: bailei@pjlab.org.cn).}
\thanks{Xiaoming Liu is with the Department of Computer Science and Engineering, Michigan State University, East Lansing, MI 48824, USA (email: liuxm@msu.edu).}
\thanks{Corresponding author: Xiaohong Liu}
}



\IEEEtitleabstractindextext{
\justifying
\begin{abstract}
The proliferation of videos generated by diffusion models has raised increasing concerns about information security, highlighting the urgent need for reliable detection of synthetic media. Existing methods primarily focus on image-level forgery detection, leaving generic video-level forgery detection largely underexplored.
To advance video forensics, we propose a consolidated multimodal detection algorithm, named MM-Det++, specifically designed for detecting diffusion-generated videos. Our approach consists of two innovative branches and a Unified Multimodal Learning (UML) module. Specifically, the Spatio-Temporal (ST) branch employs a novel Frame-Centric Vision Transformer (FC-ViT) to aggregate spatio-temporal information for detecting diffusion-generated videos, where the FC-tokens enable the capture of holistic forgery traces from each video frame. In parallel, the Multimodal (MM) branch adopts a learnable reasoning paradigm to acquire Multimodal Forgery Representation (MFR) by harnessing the powerful comprehension and reasoning capabilities of Multimodal Large Language Models (MLLMs), which discerns the forgery traces from a flexible semantic perspective. To integrate multimodal representations into a coherent space, a UML module is introduced to consolidate the generalization ability of MM-Det++. In addition, we also establish a large-scale and comprehensive Diffusion Video Forensics (DVF) dataset to advance research in video forgery detection. Extensive experiments demonstrate the superiority of MM-Det++ and highlight the effectiveness of unified multimodal forgery learning in detecting diffusion-generated videos. Code and dataset are available at \href{https://github.com/SparkleXFantasy/MM-Det-Plus}{https://github.com/SparkleXFantasy/MM-Det-Plus}.

\end{abstract}

}


\maketitle
\IEEEdisplaynontitleabstractindextext

\section{Introduction}
\IEEEPARstart{R}{ecent} years have witnessed prominent advancements in various generative diffusion methods. As one of the core techniques in the era of AI-Generated Content (AIGC), they give rise to extraordinary visually compelling content in video generation~\cite{chen2023videocrafter1, blattmann2023stable, xing2024dynamicrafter}. Advanced diffusion-based video generation approaches (\textit{e.g.}, sora, pika, \textit{etc.}) have achieved remarkable results with a wide range of visual contexts, which are often indistinguishable from real camera-captured videos. Although these latest approaches have impressed society with their versatility, the generated content poses a risk of malicious misuse, such as counterfeit faces~\cite{thies2016face2face} and falsified business content.
Therefore, this growing realism poses significant challenges to media authenticity and has raised increasing concerns in the field of digital forensics. 

\begin{figure}[t]
  \centering
  \includegraphics[width=1.0\linewidth]{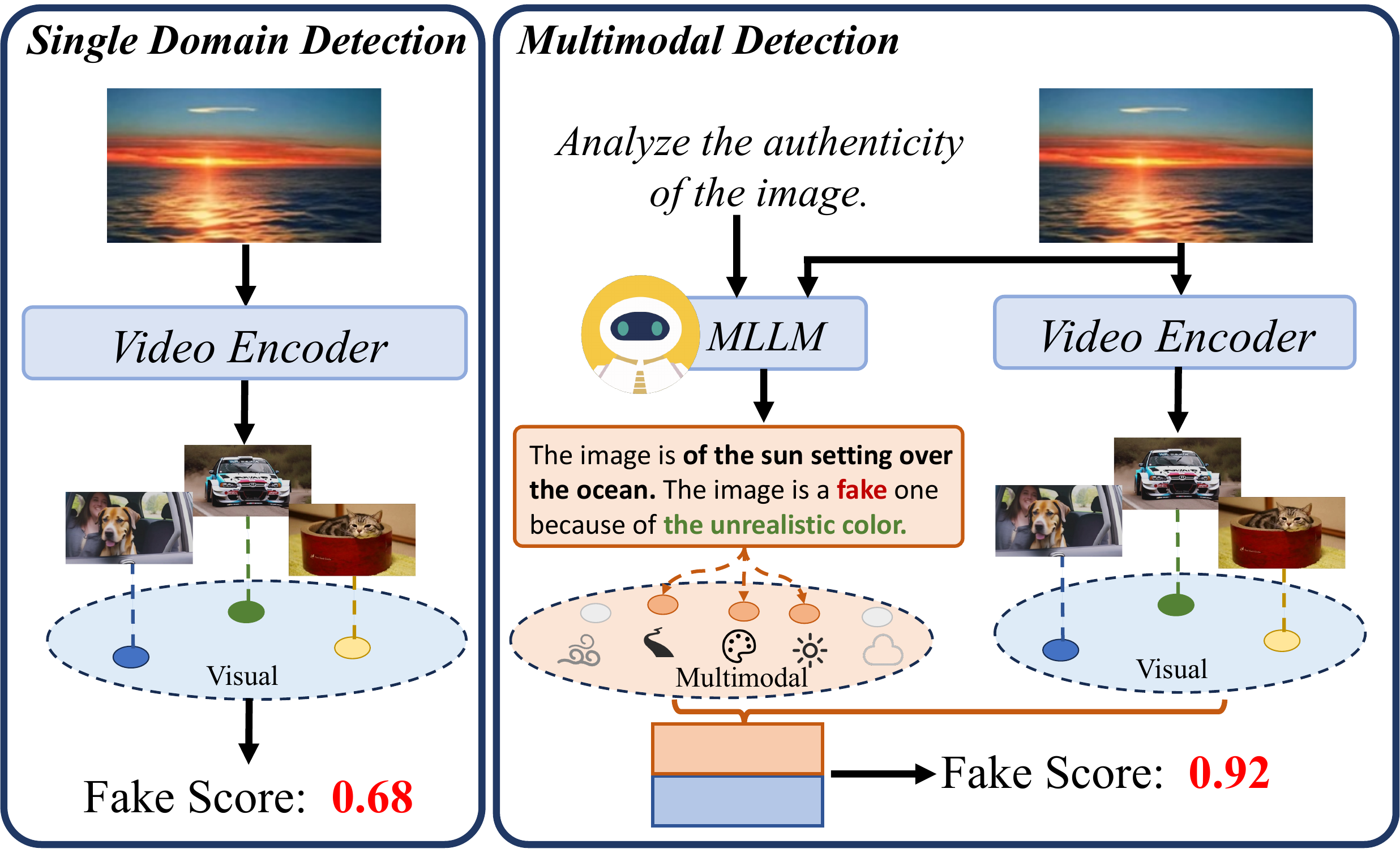}
  \caption{Comparison between single-domain and our multimodal detection frameworks. Conventional single-domain detection primarily extracts spatio-temporal traces from videos. In contrast, we propose a multimodal detection that captures forgery traces across multiple modalities. By harnessing the perceptual and reasoning capabilities of a Multimodal Large Language Model (MLLM), our framework demonstrates enhanced effectiveness in detecting the diffusion-generated videos.}
  \label{fig:thumbnail}
\end{figure}


Researchers have made significant progress in forgery detection, including image editing manipulation~\cite{chen2021image, liu2022pscc, guillaro2023trufor} and CNN-synthesized images~\cite{wang2020cnn,ricker2022towards, guo2023hierarchical}. Recently, a new research trend in detecting diffusion-generated content has emerged such as \cite{corvi2023detection} and \cite{ojha2023towards}, which focus on learning common generation artifacts and developing image-level detectors. However, since they are not specifically designed for video forensics, these methods struggle to effectively capture temporal inconsistencies, thus fall short in detecting realistic AI-generated videos with diverse manipulation traces. In addition, Multimodal Large Language Models (MLLMs) have demonstrated unprecedented comprehension and reasoning capabilities~\cite{alayrac2022flamingo, li2022blip, li2023blip, zhu2023minigpt, zhang2024llama, liu2024visual}, due to their powerful multimodal representation ability. However, such representations remain largely underexplored in the video forensics.

\begin{figure*}[t]
    \centering
    \captionsetup[subfloat]{labelfont=footnotesize,textfont=footnotesize}
    \renewcommand{\thesubfigure}{{a}}
    \subfloat[Motivation of leveraging reasoning capabilities of MLLMs in forensics.]{\includegraphics[width=0.47\linewidth]{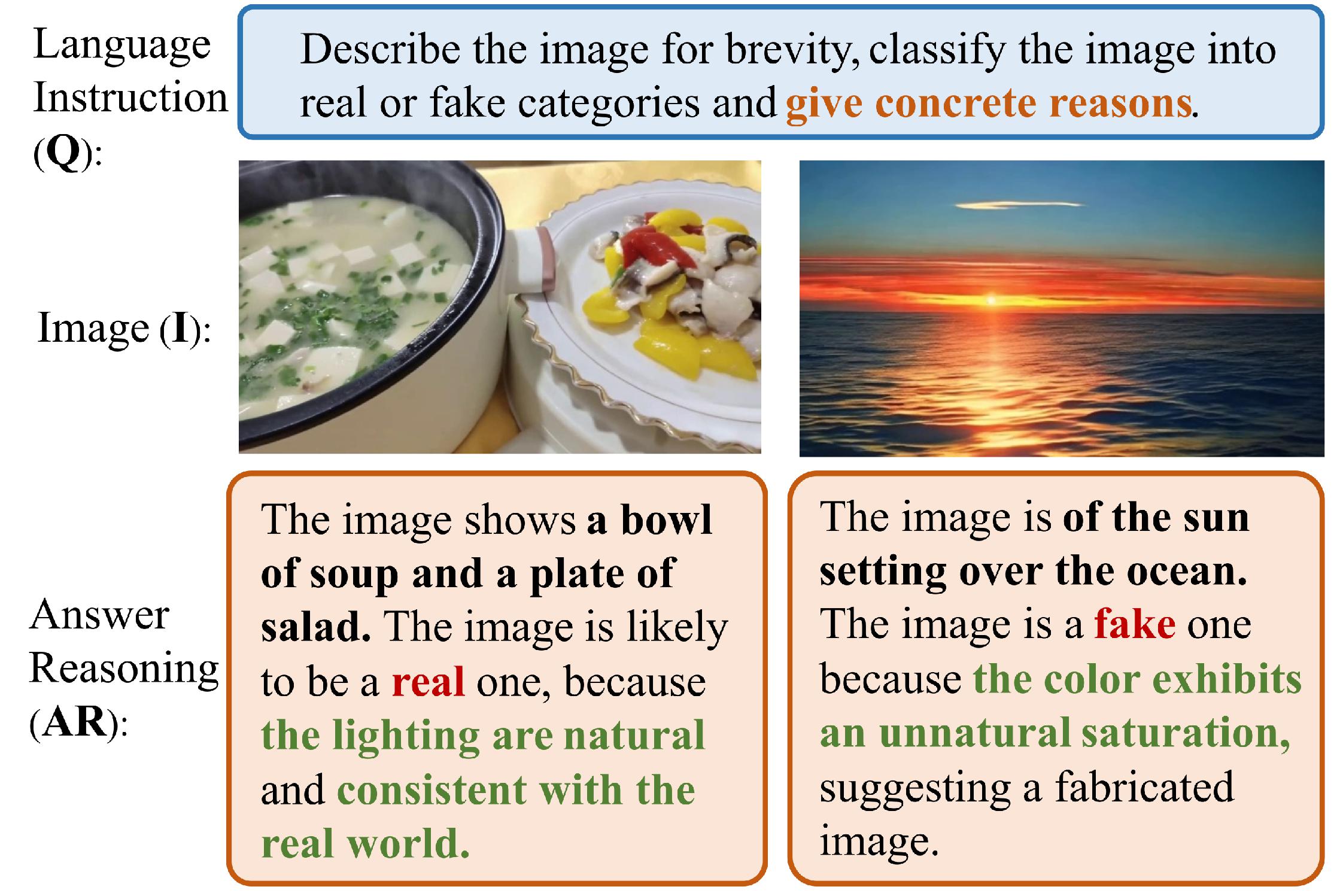}}
    \label{fig:main_graph_convs}
    \renewcommand{\thesubfigure}{{b}}
    \subfloat[The structural distinction between MM-Det and MM-Det++.]{\includegraphics[width=0.31\linewidth]{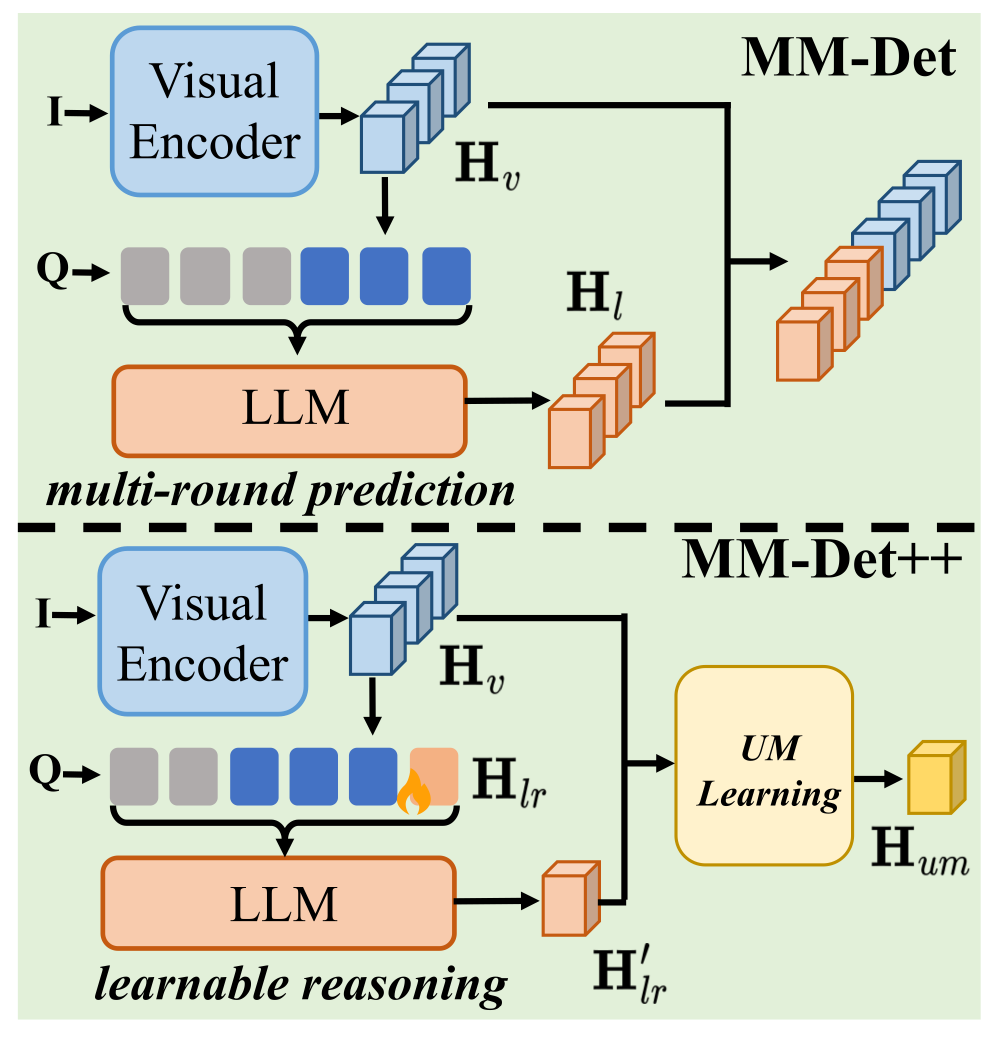}}
    \label{fig:main_graph_motivation}
    \renewcommand{\thesubfigure}{{c}}
    \subfloat[Performance of MM-Det++ over MM-Det.~\cite{song2024onlearning}]{\includegraphics[width=0.2\linewidth, height=5.9cm]{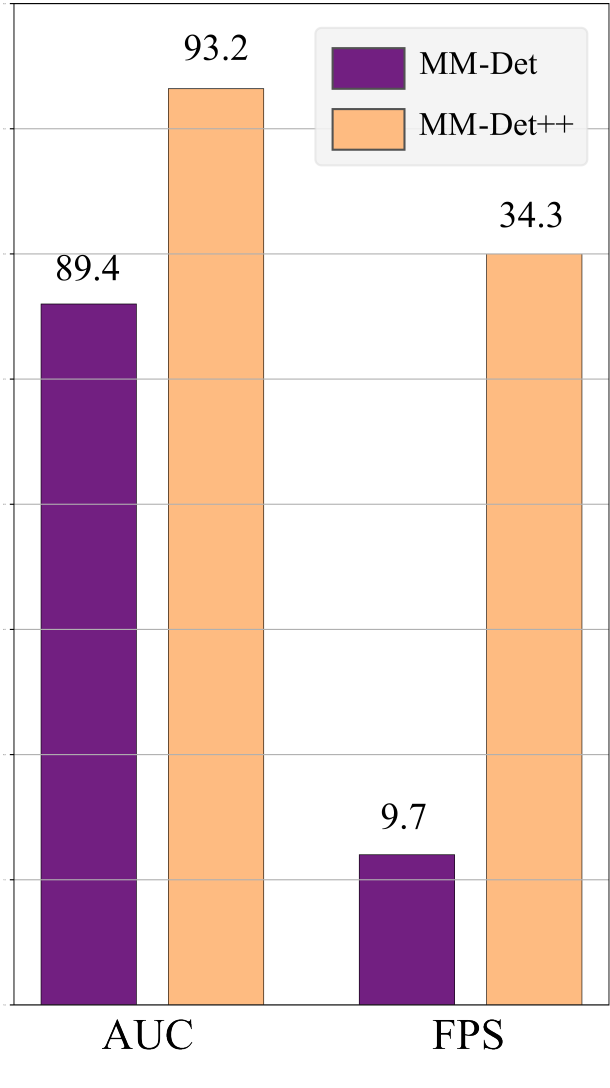}}
    \label{fig:main_graph_statistics}
    \caption{
    (a) We adopt a Multimodal Large Language Model (MLLM)~\cite{liu2024visual} to the forgery detection task. Given the language instruction $\mathbf{Q}$ and image $\mathbf{I}$ as inputs, the MLLM generates detailed and persuasive textual Answer Reasoning ($\mathbf{AR}$) to determine whether $\mathbf{I}$ is generated with AI techniques.
    (b) Comparison between the framework of MM-Det~\cite{song2024onlearning} and MM-Det++. While MM-Det incorporates visual representation $\mathbf{H}_{v}$ and reasoning representation $\mathbf{H}_{r}$ from the Large Language Model (LLM) through the multi-round prediction for forgery detection, MM-Det++ advances designs in two aspects. First, MM-Det++ replaces the time-consuming prediction with a learnable reasoning process for efficient LLM reasoning. By introducing an additional learnable reasoning token $\mathbf{H}_{lr}$ into the instruction prompt, MM-Det++ extracts reasoning-relevant information in a single prediction, significantly enhancing the efficiency. Secondly, MM-Det++ proposes a Unified Multimodal Learning (UML) module to aggregate cross-modality forgery information into a unified multimodal representation $\mathbf{H}_{um}$. 
    (c) 
    Compared with MM-Det, MM-Det++ exhibits improved efficiency and effectiveness during inference with MLLM.
    }
    \label{fig:overview}
\end{figure*}




Compared to existing single-domain approaches, we build upon the unparalleled capabilities of MLLMs to propose a novel multimodal framework for diffusion-generated video detection. Fig.~\ref{fig:thumbnail} illustrates the comparison between single-domain and our proposed multimodal detection frameworks. The motivation for leveraging the perceptual and reasoning capabilities of MLLM in forensic detection is shown in Fig.~\ref{fig:overview}\textcolor{blue}{a}. With tailored language instruction and prompt tuning, the MLLM can reason about forgery traces from a semantic perspective, providing a valuable complement to existing methods that primarily explore forgery traces from a structural perspective. 

To this end, we propose a novel multimodal detection approach, dubbed MM-Det++, which utilizes the MLLM's capabilities to counter the rapid advancements in diffusion-based video generation. MM-Det++ comprises a Multimodal (MM) branch and a Spatio-Temporal (ST) branch. The MM branch leverages the MLLM to acquire the scene-agnostic Multimodal Forgery Representation (MFR). In parallel, the ST branch captures the two prevalent manifestations of generative artifacts commonly observed in forgery videos, \textit{i.e.,} spatial distortions and temporal inconsistencies, where spatial distortions refer to anomalies within individual frames, such as unnatural textures or color discrepancies, while temporal inconsistencies involve irregularities across consecutive frames, such as unnatural motion or shape alteration. Finally, to unify the multimodal representations into a coherent space, we introduce a Unified Multimodal Learning (UML) module that facilitates effective fusion across modalities for the detection of diverse types of diffusion-generated videos.

This work is a substantial extension of our previous MM-Det~\cite{song2024onlearning}, a pioneering effort that demonstrated the potential of leveraging the extraordinary context-awareness and comprehension capabilities of MLLMs for detecting diffusion-generated videos. Nevertheless, MM-Det still faces challenges in three key areas: 1) it directly adopts multi-round MLLM-generated conversations for answer reasoning. However, as this textual representation is pretrained primarily for generic language tasks, it might be suboptimal for video forensics. 2) MM-Det simply concatenates visual and reasoning representations, which limits its ability to capture discriminative cues across modalities, thus diminishing the effectiveness in distinguishing real from generated videos;
3) MM-Det integrates a pretrained autoencoder to detect frame-level reconstruction errors in diffusion-generated content, resulting in cumbersome computational overhead and high resource consumption.

To address these challenges, MM-Det++ first leverages a learnable reasoning paradigm, where the embedding of an additional token constructs a learnable input prompt that explicitly facilitates the extraction of discriminative reasoning information from MLLMs. Then, MM-Det++ employs cross-attention to consolidate the multimodal representation aggregation. Fig.~\ref{fig:overview}\textcolor{blue}{b} demonstrates the comparison between MM-Det and MM-Det++ in acquiring the multimodal forgery representation. At last, MM-Det++ abandons the time-consuming reconstruction procedure in Frame-Centric ViT (FC-ViT). Instead, it employs a specialized attention mechanism to jointly capture spatial and temporal forgery traces, thus enhancing detection accuracy. With these improvements, MM-Det++ surpasses its predecessor in both detection effectiveness and efficiency, as illustrated in Fig.~\ref{fig:overview}\textcolor{blue}{c}. In addition, recognizing the scarcity of public datasets for diffusion-generated videos, we elaborate a comprehensive dataset, named Diffusion Video Forensics (DVF). DVF covers a diverse range of content generated by various diffusion models, ensuring rich semantics and high quality. We will continuously collect the generated videos from the latest advances, with the aim of establishing a comprehensive and evolving benchmark to facilitate the development and evaluation of open-world video forensics.



To sum up, this work substantially extends our preliminary study, MM-Det~\cite{song2024onlearning}, published at The Thirty-Eighth Annual Conference on Neural Information Processing Systems (NeurIPS'24). The main contributions are listed as follows:


$\diamond$ We propose a novel multimodal dual-branch detector called MM-Det++ that first leverages the powerful comprehension and reasoning capabilities of MLLMs for diffusion-generated video detection.

$\diamond$ To consolidate detection performance, we introduce an innovative Frame-Centric Vision Transformer (FC-ViT) to effectively aggregate spatio-temporal forgery traces, employ a learnable reasoning paradigm to derive the Multimodal Forgery Representation (MFR) from MLLMs, and design a Unified Multimodal Learning (UML) module to integrate multimodal representations into a coherent space.

$\diamond$ We present the Diffusion Video Forensics (DVF) dataset, a comprehensive benchmark comprising diverse diffusion-generated videos from various advanced methods, designed to facilitate research in diffusion-generated video forensics. 

$\diamond$ Compared with state-of-the-art detectors, MM-Det++ demonstrates the superior detection performance. The promising results on unseen fake videos generated by the latest diffusion-based methods further validate the remarkable generalizable ability of MM-Det++.





\section{Related Works}
\label{sec:related_works}

\subsection{Image-Level Detector} 
Early works~\cite{on-the-detection-of-digital-face-manipulation,wang2020cnn,jeong2022bihpf,gragnaniello2021gan,yu2019attributing,tan2023learning,proactive-image-manipulation-detection} revealed that forgery traces are present in images produced by AI techniques. 
For example, generation fingerprints have been identified in content produced by Generative Adversarial Networks (GAN)~\cite{yu2019attributing,wang2020cnn} and diffusion models~\cite{corvi2023intriguing}. To detect more subtle forgery patterns, a series of studies~\cite{tan2024rethinking, guo2023hierarchical, corvi2023detection} have introduced frequency-based clues as generative artifacts. Tan \textit{et al.}~\cite{tan2024rethinking} disclosed neighbouring pixel relationships in generative content via a downsampling operation. Guo \textit{et al.}~\cite{guo2023hierarchical} developed a hierarchical taxonomy for forgery detection and attribution on both image and frequency domains. Corvi \textit{et al.}~\cite{corvi2023detection} extracted periodical patterns in spectra maps to distinguish diffusion content from real images.
Besides, some studies~\cite{ojha2023towards,cozzolino2024raising,liu2024cfpl,liu2024forgery} extract discriminative features for synthetic content from pre-trained image encoders via Contrastive Learning Image Pretraining (CLIP)~\cite{radford2021learning}. Other approaches~\cite{wang2023dire,ma2023exposing,luo2024lare} quantify reconstruction errors between an image and the diffusion-based reconstructed counterpart, disclosing the key discrepancies between authentic and diffusion-generated content.
In addition, some researchers~\cite{yan2024transcending, lin2024preserving, mandelli2022detecting, guo2025language} focus on the enhancement of latent feature spaces for more generalizable forgery detectors. For instance, Yan \textit{et al.}~\cite{yan2024transcending} introduced an augmentation method in the latent feature spaces to distill knowledge across different forgery types. Lin \textit{et al.}~\cite{lin2024preserving} proposed a fair loss function to learn domain-agnostic forgery features. Mandelli \textit{et al.}~\cite{mandelli2022detecting} justified that diversity in the training set significantly contributes to improved generalization.
Guo \textit{et al.}~\cite{guo2025language} introduced a language-guided training strategy to enhance visual forgery features.
Compared with the prior work that focused on the single visual domain, our proposed method incorporates multimodal representations that leverage both visual information and reasoning capabilities of MLLMs to achieve remarkable generalization performance.

\subsection{Video-Level Detector} 
A series of studies have explored diverse cues and representations for facial forgery detection at the video level. Early approaches~\cite{li2018ictu, haliassos2021lips} leveraged physiological signals, where methods exploited biometric patterns that are difficult to synthesize realistically. For instance, Li \textit{et al.}~\cite{li2018ictu} proposed dynamic patterns of eye-blinking as cues for detecting fake videos. Haliassos \textit{et al.}~\cite{haliassos2021lips} distinguished fake videos by analyzing the inconsistency of mouth motion. A few methods focused on locating spatial manipulation artifacts. For example, Li \textit{et al.}~\cite{li2020face} learned the subtle boundary artifacts between the original background and manipulated faces. Wodajo \textit{et al.}~\cite{wodajo2021deepfake} extracted spatial forgery traces and modeled the pixel-level correlation using CNNs. Zhao \textit{et al.}~\cite{zhao2021multi} employed a multi-attentional detector to capture local deepfake artifacts.
Meanwhile, frequency-domain analyses have emerged as an effective strategy for facial forgery detection as well.
Masi \textit{et al.}~\cite{masi2020two} developed a hybrid approach that combines forgery features in both the color space and frequency domain using a band-pass filter.
Qian \textit{et al.}~\cite{qian2020thinking} proposed a comprehensive frequency analysis method that captures both global and local forgery traces within the frequency domain. Guo \textit{et al.}~\cite{guo2025rethinking} aligned frequency-based features with the natural language domain for interpretable forgery facial detection. 
In addition, a thread of work \cite{cozzolino2021id, gu2021spatiotemporal, zheng2021exploring, wang2023altfreezing} has explored temporal inconsistencies in facial forgery videos. For example, Zheng \textit{et al.}~\cite{zheng2021exploring} leveraged spatial information and long-term temporal coherence via flexible convolutions. Wang \textit{et al.}~\cite{wang2023altfreezing} proposed a training strategy to alternately capture both spatial and temporal forgery traces.
However, these studies are confined to facial manipulation, lacking the capabilities to detect general AIGC. In contrast, we develop a general detector for diffusion-generated video content, advancing the frontier of AIGC detection.

\subsection{MLLMs for Downstream Tasks}
Multimodal Large Language Models~\cite{alayrac2022flamingo,li2023blip,liu2024improved,liu2024visual,li2023videochat,zhang2023video,su2023pandagpt} possess generalizable problem-solving abilities due to aligned knowledge from visual and textual domains, which significantly benefit various downstream applications~\cite{gu2022open,zhang2024next,wu2024next,merullo2023linearly}. For instance, 
Koh \textit{et al.}~\cite{koh2023grounding} and Lin \textit{et al.}~\cite{lin2023towards} introduced expert multimodal language models for visual-text retrieval.
Kalantidis \textit{et al.}~\cite{kalantidis2024label} improved zero-shot visual classification through visual-language models.
Phan \textit{et al.}~\cite{phan2024decomposing} leveraged medical images with prior textual knowledge to enhance pathology detection.
Wang \textit{et al.}~\cite{wang2024visionllm} and Chen \textit{et al.}~\cite{chen2024videollm} guided visual-language models to comprehend images and videos via instruction-tuning. Pang \textit{et al.}~\cite{pang2024frozen} employed frozen transformer layers from a pre-trained large language model to encode visual information for image classification tasks.
Zhou \textit{et al.}~\cite{zhou2022learning, zhou2022conditional} proposed prompt learning techniques to adapt pre-trained vision-language models for downstream visual tasks, such as visual recognition, understanding and image-text retrieval.
These applications demonstrate the powerful multimodal capabilities of MLLMs in comprehension and reasoning downstream tasks. In this paper, we extend the application of MLLMs in diffusion-generated video detection, introducing a new framework for generalizable video forensics.

\begin{figure*}[t]
  \centering
  \includegraphics[width=1.0\linewidth]{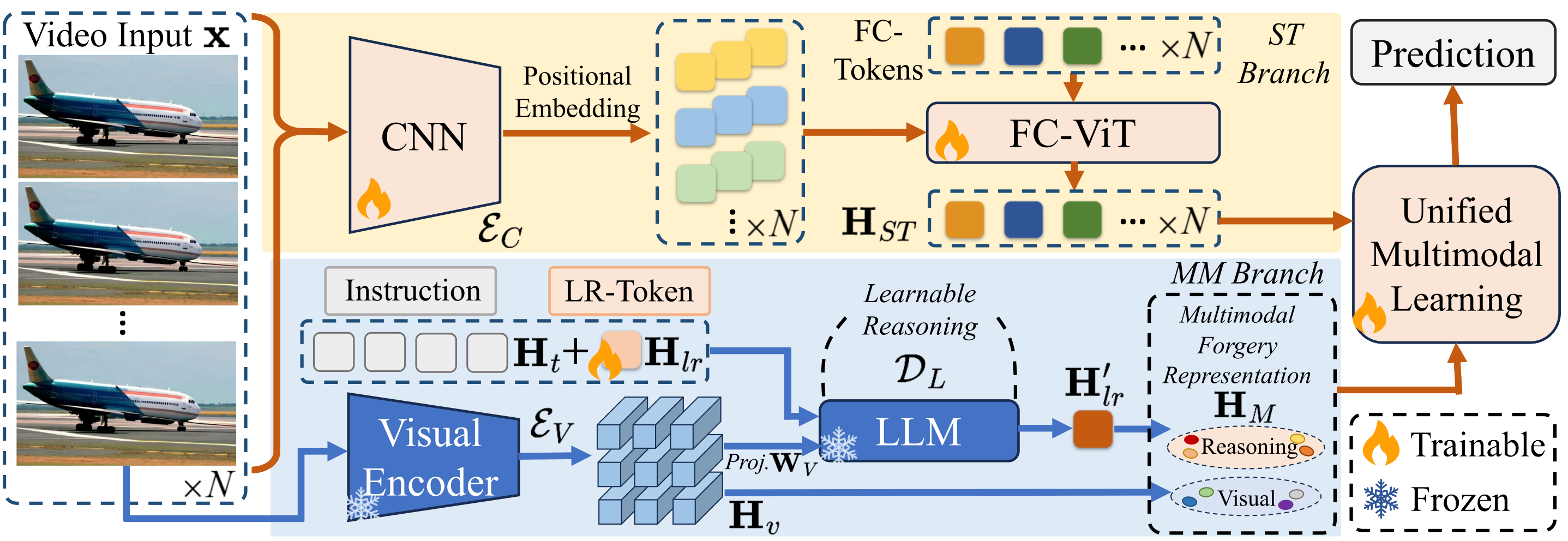}
  \caption{The overall structure of MM-Det++. It is a multimodal dual-branch detector that consists of a Spatio-Temporal (ST) branch, a Multimodal (MM) branch, and a Unified Multimodal Learning (UML) module to output the final prediction.}
  \label{fig:overview_of_main_framework}
\end{figure*}
\section{Method}
\label{sec:method}
\subsection{Overview}
The overall structure of our proposed MM-Det++ is demonstrated in Fig.~\ref{fig:overview_of_main_framework}. Given a video $\mathbf{X}$ as the input, the Spatio-Temporal (ST) branch learns spatial artifacts and temporal inconsistencies. The sequence of frames is first processed by a convolutional neural network $\mathcal{E}_{C}$, and then passed through a Frame-Centric Vision Transformer (FC-ViT) that aggregates spatio-temporal information via Frame-Centric tokens (FC-tokens), yielding the spatio-temporal forgery representation $\mathbf{H}_{ST}$. In addition, the Multimodal (MM) branch takes key frames and a tailored instruction as inputs to derive the Multimodal Forgery Representation (MFR) $\mathbf{H}_M$ by leveraging the powerful comprehension and reasoning abilities of MLLM, where $\mathbf{H}_M$ consists of the context-aggregated LR token $\mathbf{H}^{\prime}_{lr}$ and the visual tokens $\mathbf{H}{v}$. At last, a Unified Multimodal Learning (UML) module is introduced to aggregate the representations from both branches for final prediction. We illustrate the details of the ST branch, MM branch, and the UML module as follows.

\subsection{Spatio-Temporal Branch}
\label{sec:method_st_forgery}

\begin{figure}[t]
  \centering
  \includegraphics[width=0.96\linewidth]{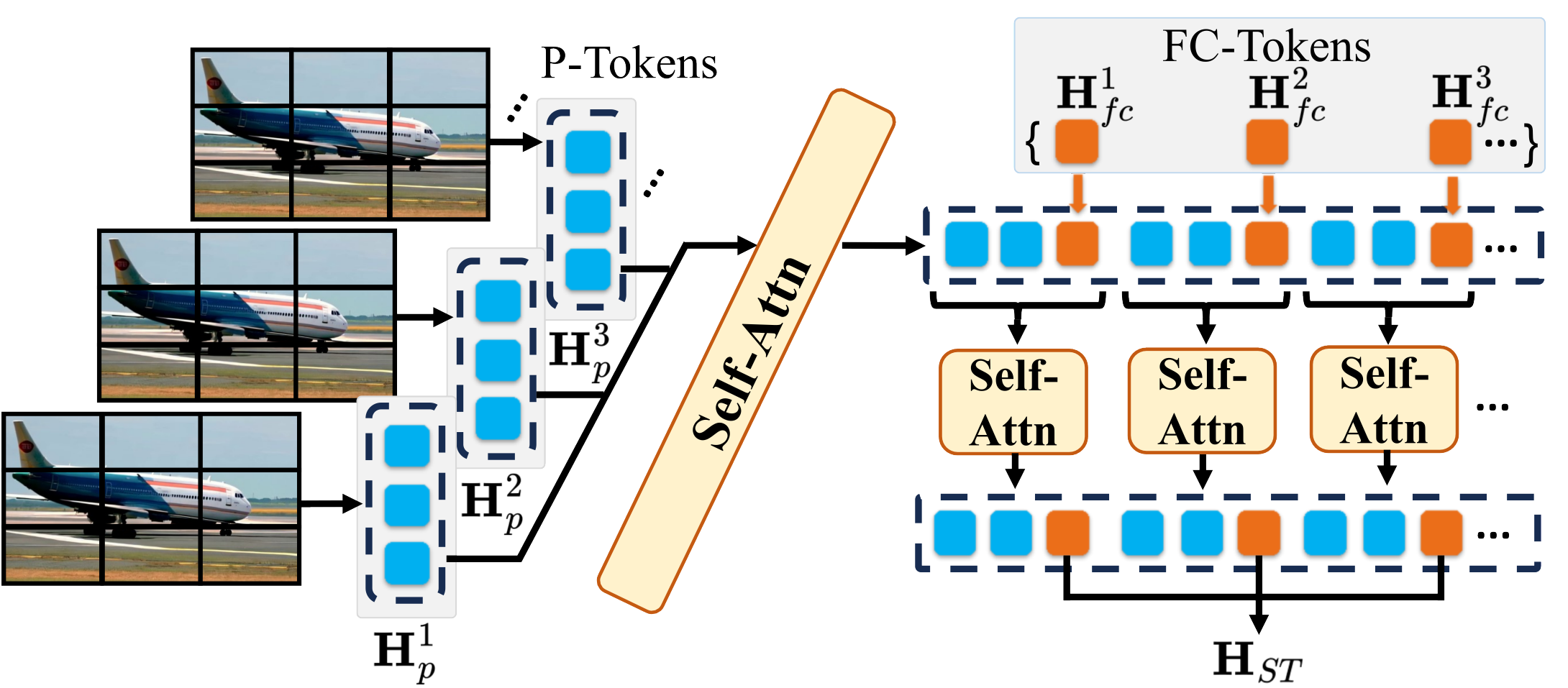}
  \caption{ 
 In FC-ViT, two self-attention operations are employed. Each input frame is first encoded and partitioned into Patch tokens (P-tokens) (\textcolor{cyan}{\rule{0.3cm}{0.25cm}}) to aggregate patch-level temporal information, while additional Frame-Centric token (FC-token) (\textcolor{orange}{\rule{0.3cm}{0.25cm}}) are introduced to aggregate frame-level spatial information, producing the spatio-temporal forgery representation $\mathbf{H}_{ST}$.
  }
  \label{fig:method_attn_st}
\end{figure}

Similar to previous works~\cite{dosovitskiy2021image, arnab2021vivit} that utilize ViT~\cite{chen2021image} to model video-level information, in the Spatio-Temporal (ST) branch, we propose a transformer-based network named FC-ViT, which consists of $12$ transformer blocks, each equipped with two consecutive self-attention operations to capture spatio-temporal forgery traces across different frames. Fig.~\ref{fig:method_attn_st} shows the detailed process within a single transformer block. Specifically, given a sequence of $N$ frames, each frame $\mathbf{x}^{i}$ where $i \in [1, N]$ is encoded by a convolutional network $\mathcal{E}_{C}$ into a feature map $\mathbf{F}^{i} = \mathcal{E}_{C}(\mathbf{x}^{i})$. The feature map $\mathbf{F}^{i}$ is then partitioned into $R$ patches as $\mathbf{F}^{i,j}$ where $j \in [1, R]$, and projected into P-tokens using a linear transformation $\mathbf{W}_{p}$ as $\mathbf{H}_{p}^{i} = \{ \mathbf{H}_{p}^{i,j}\}_{j=1}^{R} = \{ \mathbf{W}_{p}(\mathbf{F}^{i,j})\}^{R}_{j=1} \in \mathbb{R}^{R\times M}$, where $\mathbf{H}_{p}^{i}$ denotes the set of P-tokens for the $i$th frame, $\mathbf{H}_{p}^{i,j}$ represents the $j$th token within this set, and $M$ is the dimension of each P-token. 
In addition, FC-ViT introduces an additional Frame-Centric token (FC-token) for each frame ${\mathbf{x}^i}$, denoted as $\mathbf{H}_{fc}^{i} \in \mathcal{R}^{M}$.

During the aggregation of spatio-temporal forgery information, the first self-attention operation resorts to P-tokens to model the patch-level temporal dependencies that reveal the forgery traces across different frames. This operation can be formulated as 
\begin{equation}
    \mathbf{H}_{p}^{i,j} = \sum_{i=1}^{N} \sum_{j=1}^{R} \texttt{SELF-ATTN}(\mathbf{H}_{p}^{i,j}, \mathbf{H}_{p}^{k,l}),
    \label{eq:cross_frame_att}
\end{equation}
where $\texttt{SELF-ATTN}$ refers to the self-attention operation. $\mathbf{H}_{p}^{i,j} \in \mathbb{R}^{M}$ attends to the token $\mathbf{H}_{p}^{k,l} \in \mathbb{R}^{M}$ that represents the $l$th token from the $k$th frame where $ i,k \in [1, N]$, $j,l \in [1, R]$.
Subsequently, the second self-attention operation is performed between the temporal-aggregated P-tokens and the corresponding FC-token (\textit{i.e.}, $\mathbf{H}_{fc}^{i}$) within the same frame (\textit{i.e.}, $\mathbf{x}^{i}$). This operation, which effectively aggregates spatial traces with the assistance of frame-level FC-tokens, can be formulated as

\begin{equation}
    \mathbf{H}_{fc}^{i} = \sum_{j=1}^{R} \texttt{SELF-ATTN}(\mathbf{H}_{fc}^{i}, \mathbf{H}_{p}^{i,j}),
    \label{eq:in_frame_att}
\end{equation}
where the $\mathbf{H}_{fc}^{i}$ is initialized as the pretrained class token of ViT~\cite{chen2021image} in the first transformer block. Finally, FC-tokens from all frames are extracted and concatenated to form the spatio-temporal forgery representation $\mathbf{H}_{ST}$, as

\begin{equation}
    \mathbf{H}_{ST} = \texttt{CONCAT}(\mathbf{H}_{fc}^{i}), i \in [1, N].
    \label{eq:st_features}
\end{equation}

Compared with the In-and-Across Frame Attention (IAFA) proposed in MM-Det~\cite{song2024onlearning}, FC-ViT removes the time-consuming reconstruction process that reveals forgery traces produced by diffusion models. 

\subsection{Multimodal Branch}
\label{sec:method_mllm}

MM-Det++ introduces a Multimodal (MM) branch to learn representations that generalize well to unseen diffusion-generated videos, as depicted in Fig.~\ref{fig:overview_of_main_framework}. 
Specifically, the MM branch is built upon LLaVA~\cite{liu2024visual}, a representative MLLM consisting of a CLIP visual encoder $\mathcal{E}_V$ and a large language model $\mathcal{D}_L$ (\textit{i.e.}, Llama-2~\cite{touvron2023llama}).
Given a key frame $\mathbf{x}^{k}$, it is fed to $\mathcal{E}_V$ to obtain the visual representation $\mathbf{H}_{v} = \mathcal{E}_V(\mathbf{x}^k) \in \mathbb{R}^{L \times S}$, where $L$ is the length of the visual sequence and $S$ is the dimension of the visual space. The representation $\mathbf{H}_{v}$ contains rich semantics and shows impressive generalization ability and robustness in forgery detection tasks~\cite{ojha2023towards, sha2023fake, cozzolino2024raising}.

\begin{figure}[t]
    \footnotesize
    \centering
    \includegraphics[width=1.\linewidth]{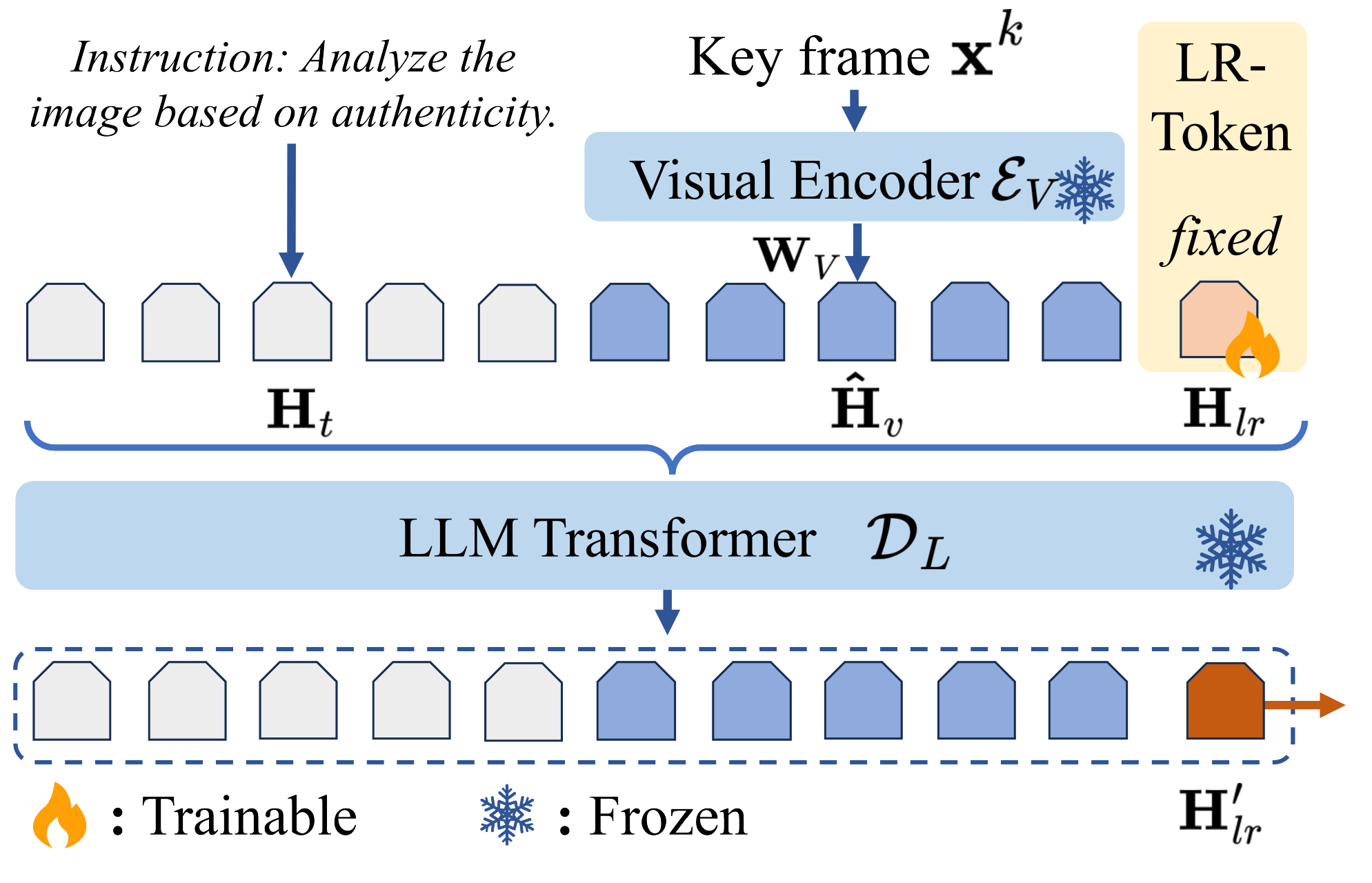}
    \caption{The learnable reasoning process in the MM branch of MM-Det++. 
    We introduce a Learnable Reasoning token (LR-Token) to capture reasoning representations from the MLLM.
    Given a key frame $\mathbf{x}^{k}$, the embedding of textual tokens $\mathbf{H}_{t}$, the aligned visual tokens $\mathbf{\hat{H}}_{v}$, and the LR-token $\mathbf{H}_{lr}$ (fixed during inference) are concatenated and fed into a frozen LLM transformer.
    At last, the context-aggregated LR-token $\mathbf{H}^{\prime}_{lr}$ is produced from $\mathcal{D}_{L}$ to facilitate reasoning over the diffusion-generated videos.}
    \label{fig:method_learnable_reasoning}
\end{figure}

We propose a learnable reasoning paradigm to capture reasoning representations from MLLM by introducing a Learnable Reasoning token (LR-token), which aggregates the context information from all engaged textual and visual tokens to learn authenticity-oriented reasoning representations. The detailed structure is shown in Fig.~\ref{fig:method_learnable_reasoning}. Specifically, a given textual instruction is tokenized into a sequence of textual tokens, denoted as $\mathbf{H}_{t} \in \mathbb{R}^{O \times D}$, where $O$ stands for the sequence length and $D$ represents the dimension of the textual embedding. For clarity, we omit the superscript $k$ of all tokens. The visual tokens $\mathbf{H}_{v} \in \mathbb{R}^{L \times S}$ is aligned with the textual space into $\mathbf{\hat{H}}_{v}$, defined as $\mathbf{\hat{H}}_{v} = \mathbf{W}_{V}(\mathbf{H}_{v}) \in \mathbb{R}^{L \times D}$, where $\mathbf{W}_{V}$ refers to the projector. The LR-token is first randomly-initialized within the same textual space, denoted as $\mathbf{H}_{lr} \in \mathbb{R}^{D}$. Subsequently, textual tokens $\mathbf{H}_{t}$, the aligned visual tokens $\mathbf{\hat{H}_{v}}$, and the LR-token $\mathbf{H}_{lr}$ are concatenated as the input of $\mathcal{D}_{L}$ for authenticity reasoning. At last, the context-aggregated LR-token $\mathbf{H}^{\prime}_{lr}$ is generated by $\mathcal{D}_{L}$, which serves as the overall reasoning representations. This efficient design also avoids the need to involve all generated tokens for subsequent computation. The overall process can be formulated as 

\begin{equation}
    \mathbf{H}^{\prime}_{lr} = \mathcal{D}_{L}(\texttt{CONCAT}(\mathbf{H}_{t}, \mathbf{\hat{H}}_{v}, \mathbf{H}_{lr})),
\end{equation}
where $\texttt{CONCAT}$ stands for the concatenation operation. 


Furthermore, we argue that $\mathbf{H}^{\prime}_{lr}$ primarily captures reasoning representations from MLLM, thus may fall short in retaining visual details. To address this issue, the final multimodal forgery representation $\mathbf{H}_{M}$ is constituted by $\mathbf{H}^{\prime}_{lr}$ and $\mathbf{H}_{v}$, which will be integrated into a cross-modal representation $\mathbf{H}_{c}$ in the UML module to facilitate mutual enhancement and semantic alignment between visual and reasoning spaces. Different from MM-Det~\cite{song2024onlearning} that depends on multi-round prediction for authenticity reasoning, MM-Det++ introduces the LR-token to acquire reasoning representations in a single pass across diverse visual scenarios.

\subsection{Unified Multimodal Learning}
\label{sec:method_um_learning}

\begin{figure}[t]
  \footnotesize
  \centering
  \includegraphics[width=1.\linewidth]{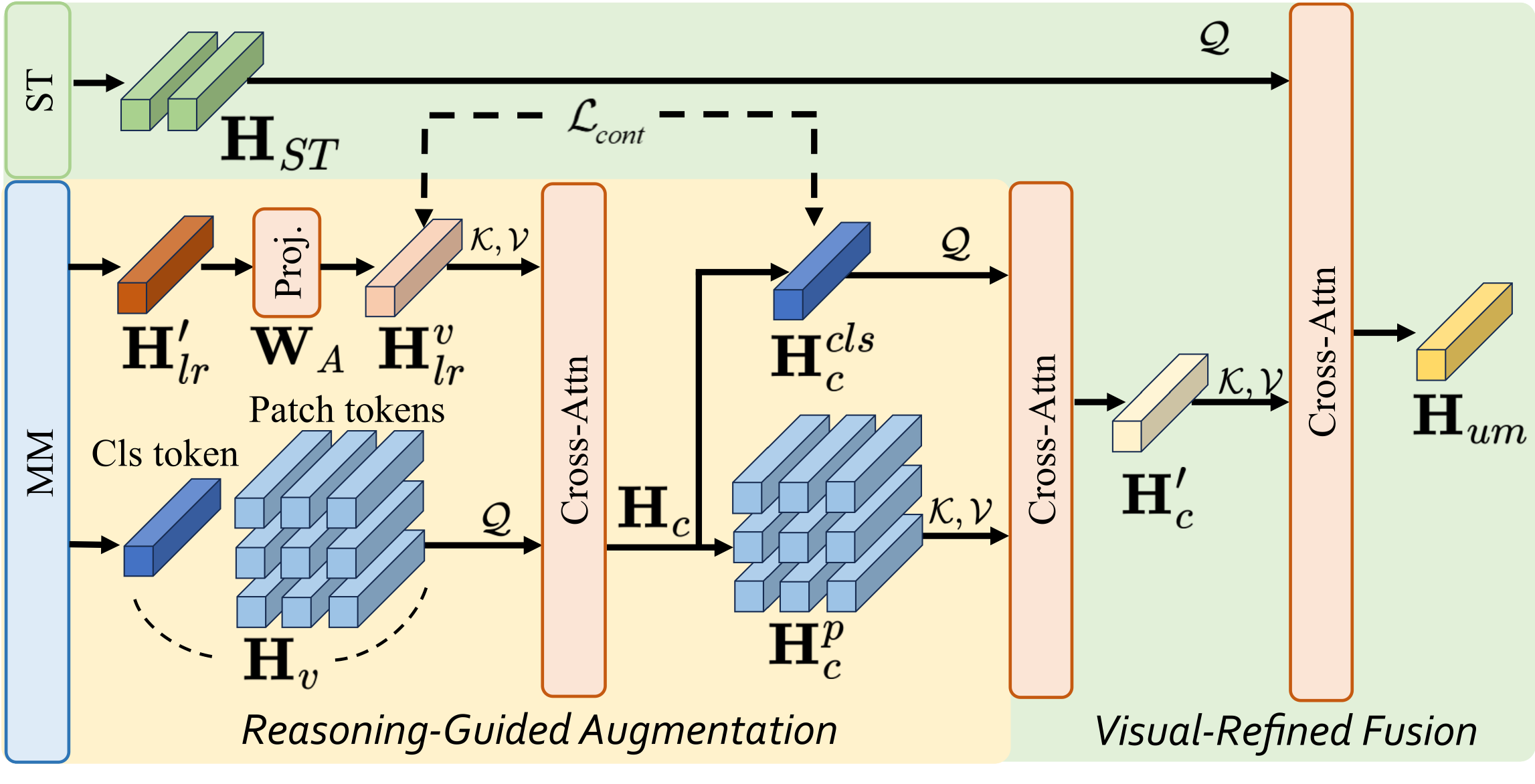}
  \caption{
    The structure of Unified Multimodal Learning (UML). Representations from multiple modalities $\mathbf{H}_{ST}$, $\mathbf{H}_{lr}^{\prime}$, and $\mathbf{H}_{v}$ are integrated and fused into a unified multimodal representation $\mathbf{H}_{um}$. We introduce reasoning-guided augmentation as well as visual-refined fusion, enabling mutual enhancement between visual and reasoning domains.
}
  \label{fig:method_um_learning}
\end{figure}

MM-Det++ proposes a Unified Multimodal Learning module to seamlessly integrate multiple modalities, as depicted in Fig.~\ref{fig:method_um_learning}. It leverages both visual and reasoning representations for mutual enhancement, yielding a unified multimodal representation $\mathbf{H}_{um}$ for forgery video detection.

Specifically, given the reasoning representation $\mathbf{H}^{\prime}_{lr} \in \mathbb{R}^{D}$, the visual representation $\mathbf{H}_{v} \in \mathbb{R}^{L \times S}$ as well as the spatio-temporal forgery representation $\mathbf{H}_{ST} \in \mathbb{R}^{N \times M}$ from multiple modalities, we first introduce a reasoning-guided augmentation to enrich the pretrained CLIP visual domain with MLLMs' reasoning. 
The reasoning representation $\mathbf{H}_{lr}^{\prime}$ is projected via $\mathbf{W}_{A}$ into the visual space as $\mathbf{H}_{lr}^{v} \in \mathbb{R}^{S}$, then fused with $\mathbf{H}_{v}$ into a cross-modal representation $\mathbf{H}_{c}$ via a cross-attention layer \texttt{CROSS-ATTN} as 

\begin{equation}
    \mathbf{H}_{c} = \texttt{CROSS-ATTN}(\mathbf{H}_{v}, \mathbf{H}_{lr}^v).
\end{equation}
Following the architectural design of CLIP, $\mathbf{H}_{c}$ is then divided into the class token $\mathbf{H}_{c}^{cls}$ that contains spatial information at the frame level and the patch-wise tokens $\mathbf{H}_{c}^{p}$ that contains semantic information at the patch level. This devision can be formulated as $\mathbf{H}_{c} = \{\mathbf{H}_{c}^{cls}, \mathbf{H}_{c}^{p}\}$. We employ contrastive learning~\cite{radford2021learning} between $\mathbf{H}_{lr}^{v}$ and $\mathbf{H}_{c}^{cls}$ to align the CLIP space with the reasoning space, which leverages the reasoning capabilities of MLLM to guide the forgery semantic learning.
The contrastive loss can be formulated as
\begin{equation}
    \mathcal{L}_{cont} = - \frac{1}{2B}\sum^{2B}_{i=1}\log(\frac{\exp(sim(z_{i}, z_{i}^{+})/\tau)}{\sum^{2B}_{j=1}\exp(sim(z_{i}, z_{j})/\tau)}),
    \label{eq:um_contrastive_loss}
\end{equation}
    where $z_{i}$ stands for the $i$th representation in the sorted set of $\mathbf{H}_{c}^{cls}$ and $\mathbf{H}_{lr}^{v}$, and $z_{i}^{+}$ denotes its corresponding positive sample, $sim$ represents the cosine similarity function, $B$ denotes the batch size, which is set to $6$, and $\tau$ is a hyperparameter fixed at $0.5$.




After reasoning-guided augmentation, $\mathbf{H}_{c}^{cls}$ is further enhanced by a visual-refined fusion, where patch-level spatial information $\mathbf{H}_{c}^{p}$ and video-level temporal information $\mathbf{H}_{ST}$ are incorporated via cross attention layers to embed multi-level visual forgery traces into $\mathbf{H}_{c}^{cls}$, yielding a unified multimodal representation $\mathbf{H}_{um}$ as the final output. This procedure is formulated as

\begin{equation}
    \mathbf{H}_{c}^{\prime} = \texttt{CROSS-ATTN}(\mathbf{H}^{cls}_{c}, \mathbf{H}^{p}_{c}),
\end{equation}
\begin{equation}
    \mathbf{H}_{um} = \texttt{CROSS-ATTN}(\mathbf{H}_{ST}, \mathbf{H}_{c}^{\prime}).
\end{equation}
The resulting $\mathbf{H}_{um}$ represents an effective integration of MLLM reasoning, patch-level spatial details, and video-level temporal cues, serving as a unified representation for diffusion-generated video detection.

\begin{figure}[t]
\centering
\includegraphics[width=0.96\linewidth]{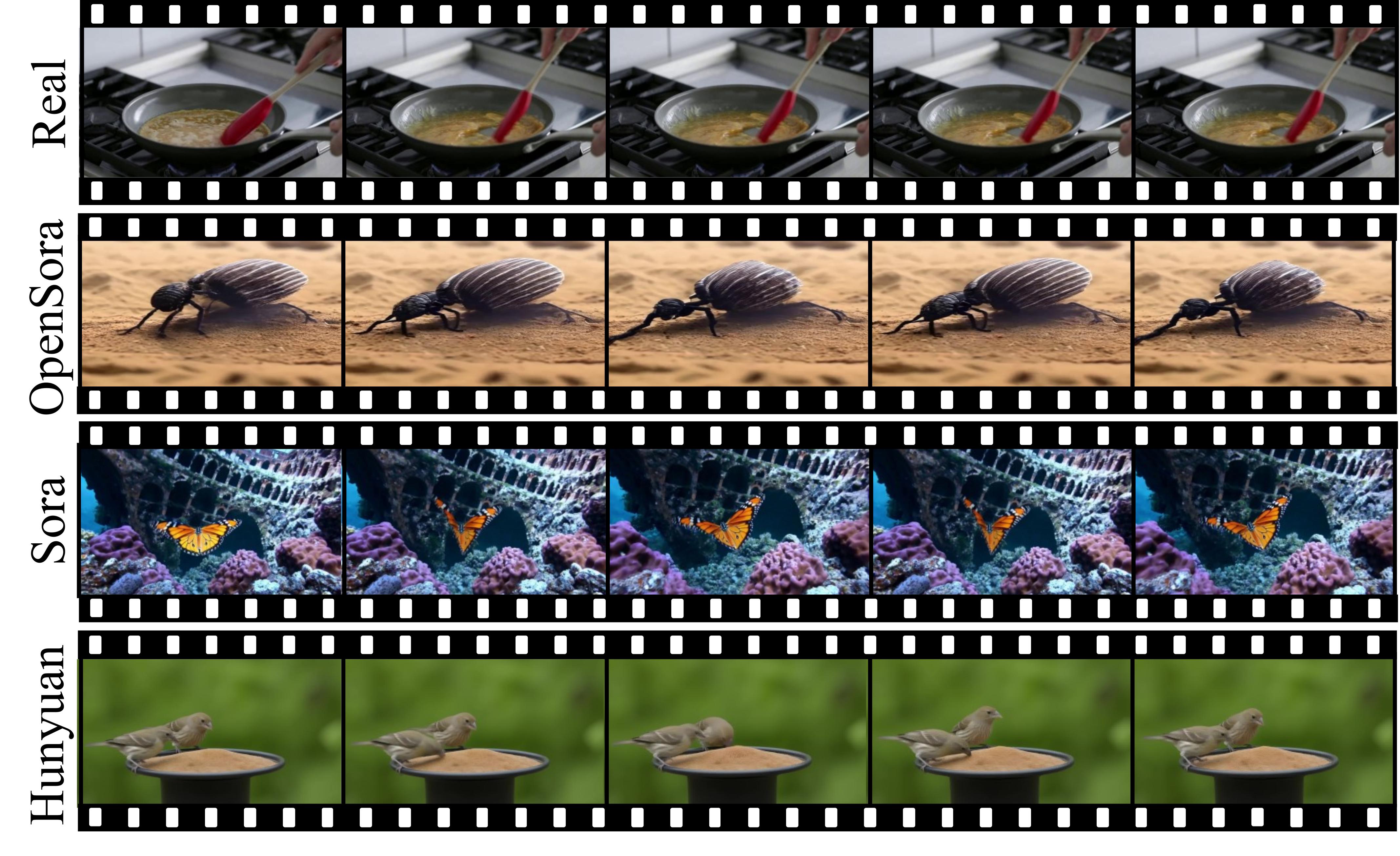}
\caption{Sampled videos from DVF dataset.}
\label{fig:overview_of_dvf}
\end{figure}

\section{Diffusion Video Forensics}

To facilitate the study of diffusion video detection, we establish a comprehensive dataset named Diffusion Video Forensics (DVF). Some sampled videos are shown in Fig.~\ref{fig:overview_of_dvf}. DVF contains $10$ diffusion generative methods that include both proprietary methods (\textit{i.e.} Sora~\cite{Sora}, Pika~\cite{Pika}, and Stable Video~\cite{StableVideo}) and open-sourced methods (\textit{i.e.} VideoCrafter~\cite{chen2023videocrafter1}, SVD~\cite{blattmann2023stable}, Zeroscope~\cite{Zeroscope}, OpenSora~\cite{Opensora}, LDM~\cite{rombach2022high}, Hunyuan~\cite{kong2024hunyuanvideo}, CogVideoX~\cite{yang2025cogvideox}). Except for SVD~\cite{blattmann2023stable}, which is an image-to-video method, all other aforementioned methods are text-to-video.

\begin{figure*}[t]
\centering
\captionsetup[subfloat]{labelfont=footnotesize,textfont=footnotesize}
\renewcommand{\thesubfigure}{{a}}
\subfloat[Generation pipeline of DVF]{\includegraphics[width=0.43\linewidth]{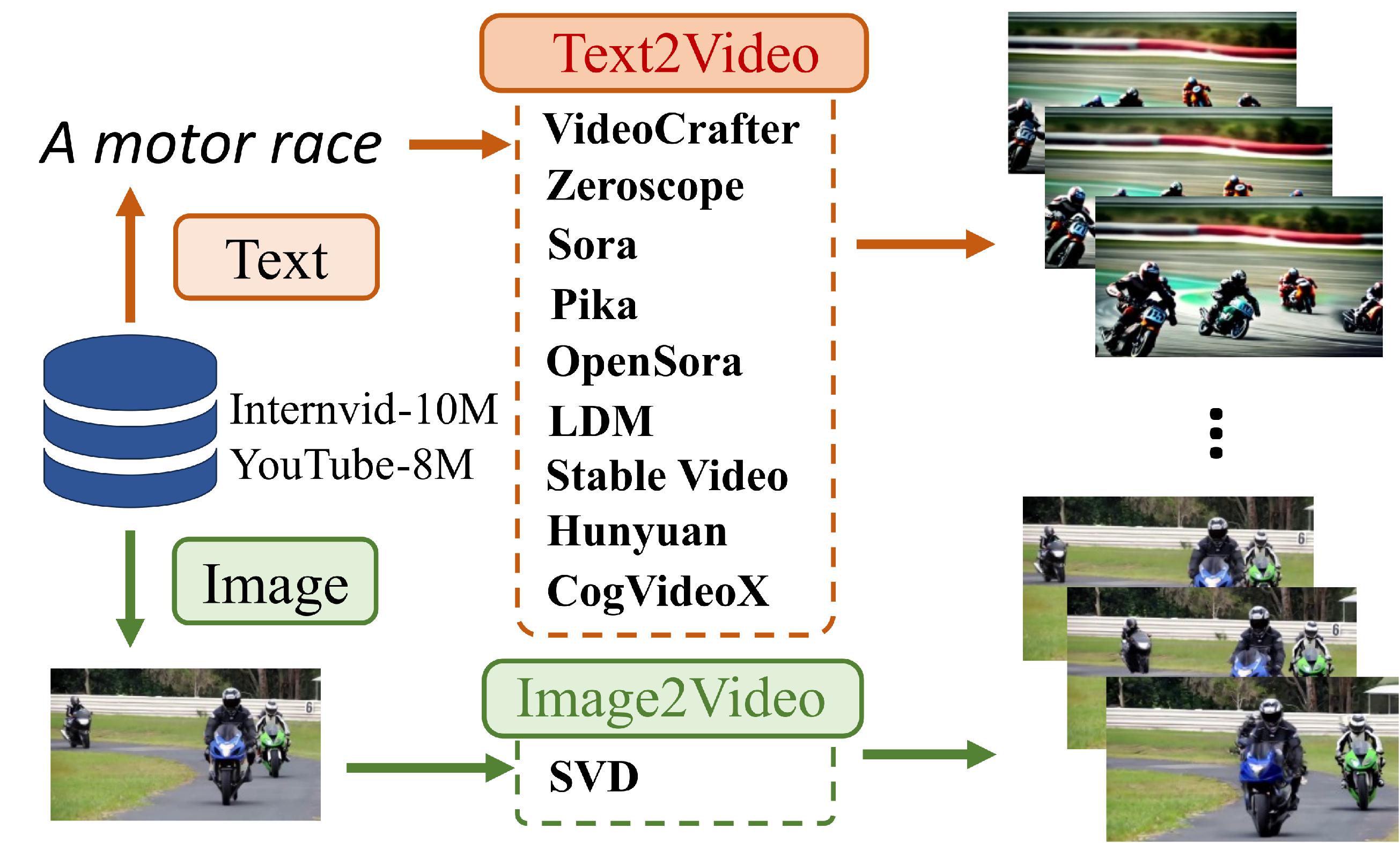}}
\label{fig:overview_dvf_pipeline}
\renewcommand{\thesubfigure}{{b}}
\subfloat[Durations and resolutions in DVF]{\includegraphics[width=0.27\linewidth]{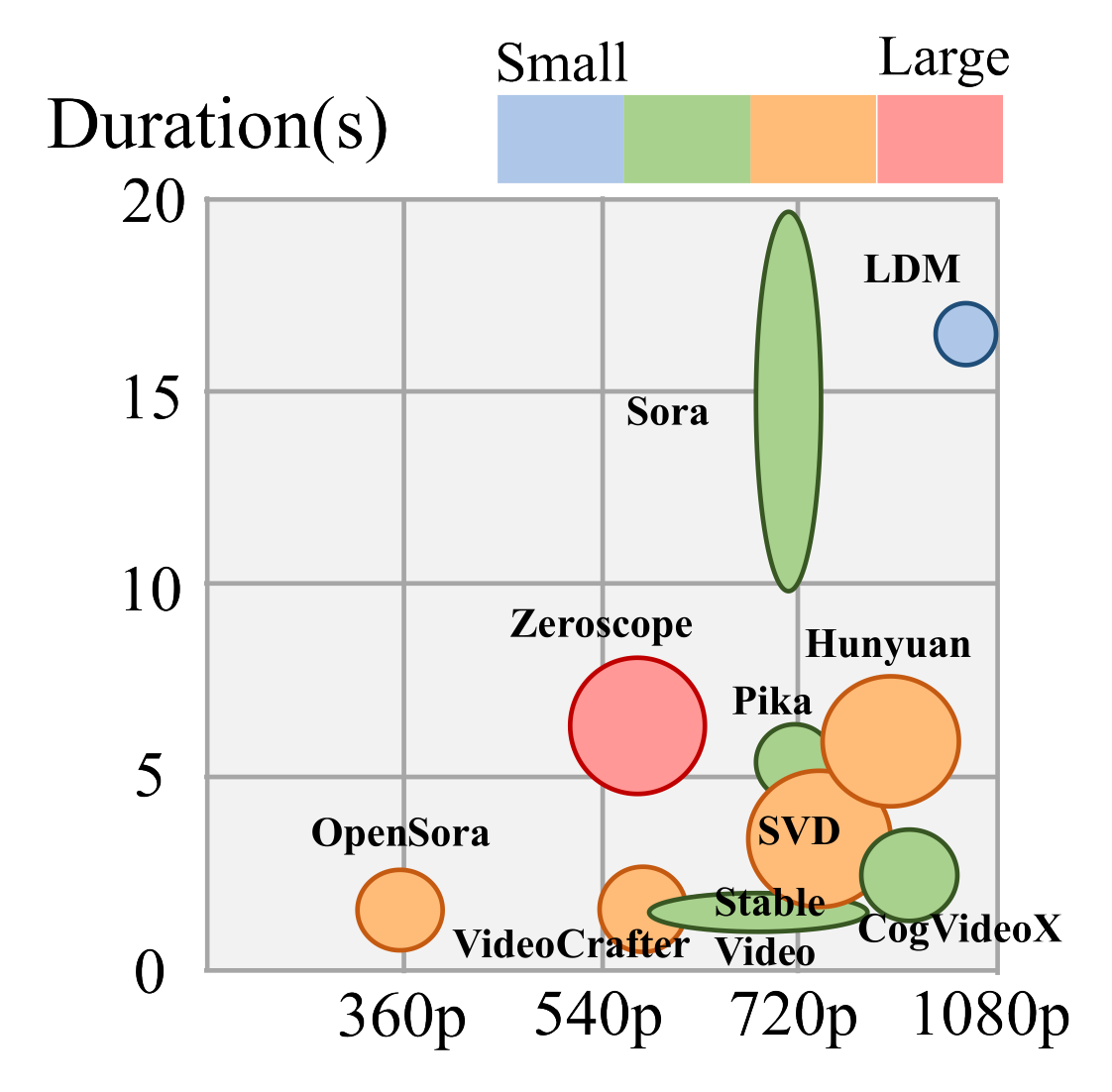}}
\label{fig:overview_dvf_metadata}
\renewcommand{\thesubfigure}{{c}}
\subfloat[Video and frame numbers in DVF]{\includegraphics[width=0.28\linewidth]{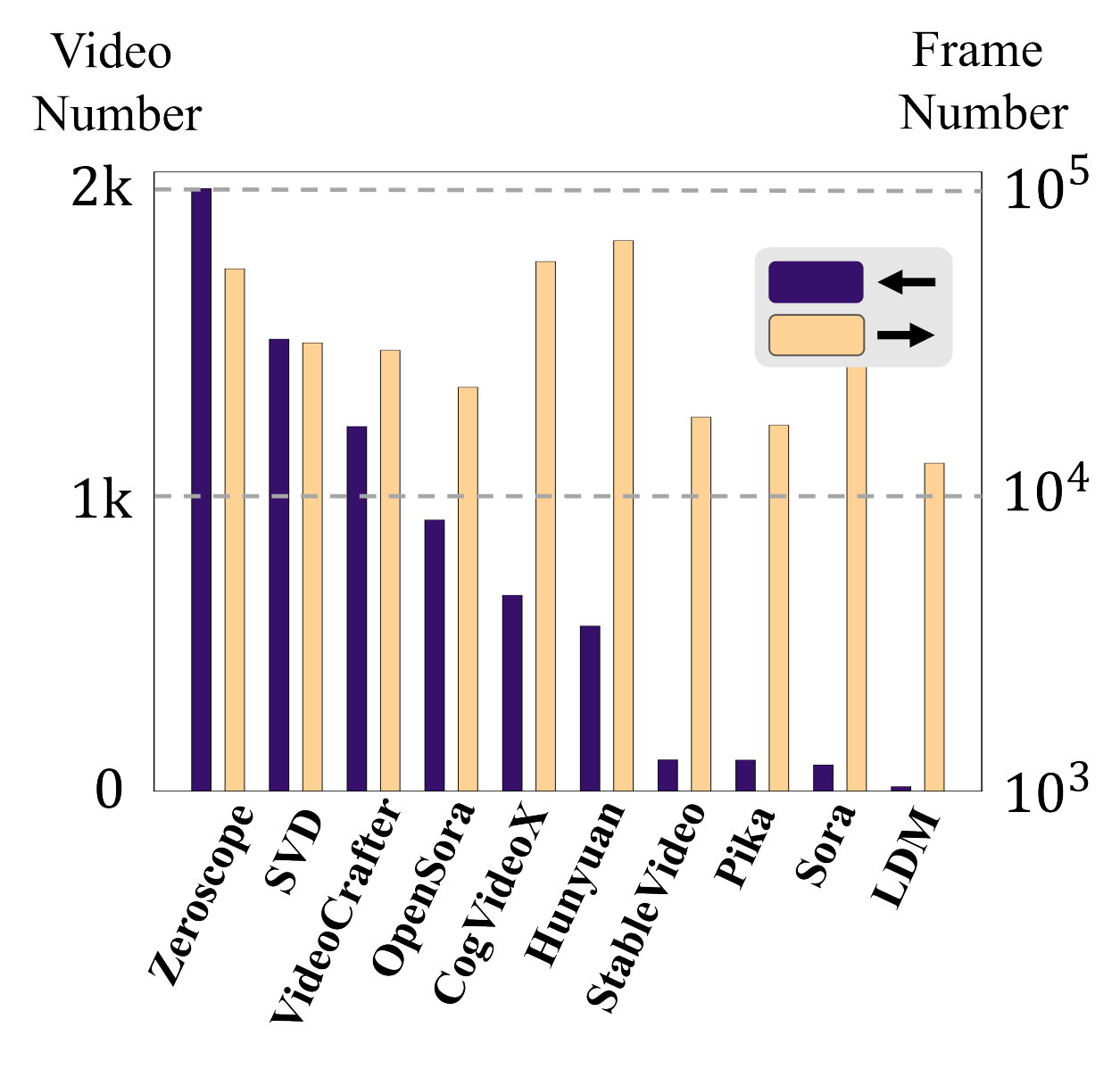}}
\label{fig:overview_dvf_number}
\caption{The overview of DVF dataset.
(a) The pipeline of forgery video generation and collection. Real frames and captions are sampled from Internvid-10M~\cite{wang2024internvid} and Youtube-8M~\cite{abu2016youtube} for text-to-video and image-to-video generation. 
(b) DVF contains videos of varying resolutions and durations, generated by multiple methods in diverse quantities. (c) The scale of each video dataset in DVF, measured by the frame and video numbers.
}
\label{fig:text_image_pair_dataset}
\end{figure*}

To prevent the diffusion-generated video detector from distinguishing forgery videos from real ones based on semantic differences, we intentionally align the semantic content between them. This design encourages the detector to exploit critical forgery traces rather than semantic cues introduced by the dataset bias, enabling an accurate assessment of forensic generalization in practical applications. To achieve this, we start from two real video datasets, Internvid-10M~\cite{wang2024internvid} and Youtube-8M~\cite{abu2016youtube}. 
For open-sourced methods, we randomly select one frame from the real video as the initial frame for image-to-video generation and leverage the paired caption for text-to-video generation, as shown in Fig.~\ref{fig:text_image_pair_dataset}\textcolor{blue}{a}. For proprietary methods, fake videos are collected directly from their official websites. We employ Video-LLaMA~\cite{zhang2023video} to generate textual descriptions of the fake videos and retrieve their real counterparts with the most semantically similar captions, measured by the similarity scores~\cite{reimers2019sentence}.

After data cleaning with diversity guaranteed, the DVF dataset has $7.1$k fake videos and $7.1$k real videos in total. As illustrated in Fig.~\ref{fig:text_image_pair_dataset}\textcolor{blue}{b}, this dataset contains diverse diffusion-generated videos from representative methods with various resolutions from $360$p to $1080$p and durations from $1$s to $20$s. Besides, the number of generated videos by different methods varies between $0.1$k to $2$k with the corresponding frames ranging from $5.5$k to $65$k in Fig.~\ref{fig:text_image_pair_dataset}\textcolor{blue}{c}.

\section{Training Strategy}
\label{sec:train_inference}

This section details our two-stage training strategy, in which we first finetune the LLM in MM branch via instruction tuning~\cite{liu2024visual} and then optimize the framework in an end-to-end manner.

\subsection{MM Branch Instruction Tuning} 
Following the visual instruction tuning~\cite{liu2024visual}, we first adapt LLaVA to the forgery detection task with LoRA~\cite{hu2022lora} to improve its reasoning ability. To this end, we construct a large image-text paired dataset for video forensics, where the ground-truth descriptions are obtained with the aid of Gemini~\cite{team2023gemini} (as detailed in Sec.~\ref{sec:expt_implementation_details}).
Subsequently, we design multi-round conversations to guide the MLLM in perceiving and identifying the authenticity of input images. It is worth noting that while multi-round conversations are employed during finetuning to strengthen the MLLM's reasoning ability in forensics, inference requires only a single pass. We finetune the projection layers and the LLM in LLaVA, leaving the visual encoder unchanged. The objective function is defined as the autoregressive loss over the answer tokens generated by the MLLM:


\begin{equation}
    \mathcal{L}(\theta_1) = - \sum_{t=1}^T log(p_{\theta_1}(s^t | s^{i < t})),
\end{equation}
where $s^t$ and $s^{i < t}$ refer to the $t$th token prediction and all tokens before $t$th token. $T$ and $\theta_1$ refer to the length of prediction tokens and the trainable parameters in the MLLM, respectively.

\subsection{End-to-End Training}
We use the finetuned LLaVA as the core component of the MM branch. Then, our MM-Det++ is trained in an end-to-end manner, where all parameters in the MM branch, except the learnable reasoning token, are frozen to ensure the acquisition of feasible reasoning.
Specifically, we represent the final prediction probability of MM-Det++ and the corresponding ground truth as $\hat{y}$ and $y$, respectively. The cross-entropy loss $\mathcal{L}_{ce}$ can be formulated as:

\begin{equation}
    \mathcal{L}_{ce} = -(y\log \hat{y} + (1 - y)\log(1 - \hat{y})).
    \label{eq:cross_entropy_loss}
\end{equation}

The model is jointly optimized with the contrastive loss in Eq.~\ref{eq:um_contrastive_loss} and the cross-entropy loss in Eq.~\ref{eq:cross_entropy_loss} as
\begin{equation}
    \mathcal{L}(\theta_{2}) = \lambda * \mathcal{L}_{cont} + \mathcal{L}_{ce},
    \label{eq:total_loss}
\end{equation}
where $\theta_2$ refers to trainable parameters in this end-to-end training stage and $\lambda$ refers to the balance weight of $\mathcal{L}_{cont}$, which is set to $1$ as default.
\section{Experiments}
\label{sec_exp}
This section presents a comprehensive evaluation of our proposed method. We first describe the experimental setup and implementation details in Sec.~\ref{sec:expt_setup} and Sec.~\ref{sec:expt_implementation_details}. Subsequently, we employ our elaborately constructed semantic-independent DVF dataset to compare the proposed MM-Det++ with other SOTAs in Sec.~\ref{sec:expt_main}, and conduct thorough ablation studies in Sec.~\ref{sec:expt_abla}, where the contribution of each component in MM-Det++ is validated. Sec.~\ref{sec:expt_mmfr_analysis} further visualizes the spatio-temporal forgery representation and multimodal forgery representation. In addition, Sec.~\ref{sec:expt:mllm_reasoning}, Sec.~\ref{sec:expt_lr_umlearning}, Sec.~\ref{sec:expt_st_t}, and Sec.~\ref{sec:expt_robustness} provide extensive analyses of the reasoning representation, unified multimodal learning, ST branch, and the robustness against various attacks, respectively.


\subsection{Experimental Setup}
\label{sec:expt_setup}
We employ the curated DVF as the benchmark for our video forensics experiments. 
For a fair comparison, we select the following $9$ representative methods as 
baselines. CNNDet~\cite{wang2020cnn} applies
a ResNet~\cite{he2016deep} as the backbone for image forgery detection. Uni-FD~\cite{ojha2023towards} leverages a pre-trained CLIP~\cite{radford2021learning} as a training-free feature space for detecting forgery content. HiFi-Net~\cite{guo2023hierarchical}, F3Net~\cite{qian2020thinking}, and NPR~\cite{tan2024rethinking} exploit frequency traces left in generated content.  DIRE~\cite{wang2023dire} identifies diffusion-generated images via a reconstruction process based on DDIM~\cite{song2021denoising}. TALL~\cite{xu2023tall} proposes a thumbnail layout strategy to efficiently preserve spatiotemporal information for deepfake video detection. TS2-Net~\cite{liu2022ts2} designs a token shift and selection mechanism for video representation learning. We compare its attention mechanism with our FC-ViT in terms of capturing forgery traces and spatial inconsistencies. In addition, our precedent work MM-Det~\cite{song2024onlearning}, the pioneer that leverages the comprehensive reasoning ability of MLLM for video forensics, is also compared. Since the generalization ability is one of the most critical factors for forensics, we decide to only use all $1.2$k fake videos from VideoCrafter~\cite{chen2023videocrafter1} and the corresponding $1.2$k real videos in DVF as the training set to train all compared methods with their publicly available source codes, while reserving the remaining videos ($5.9$k fake and $5.9$k real videos) from diverse generation methods as the test sets for cross-dataset validation. 

\begin{figure}[t]
    \centering
    \includegraphics[width=1.\linewidth]{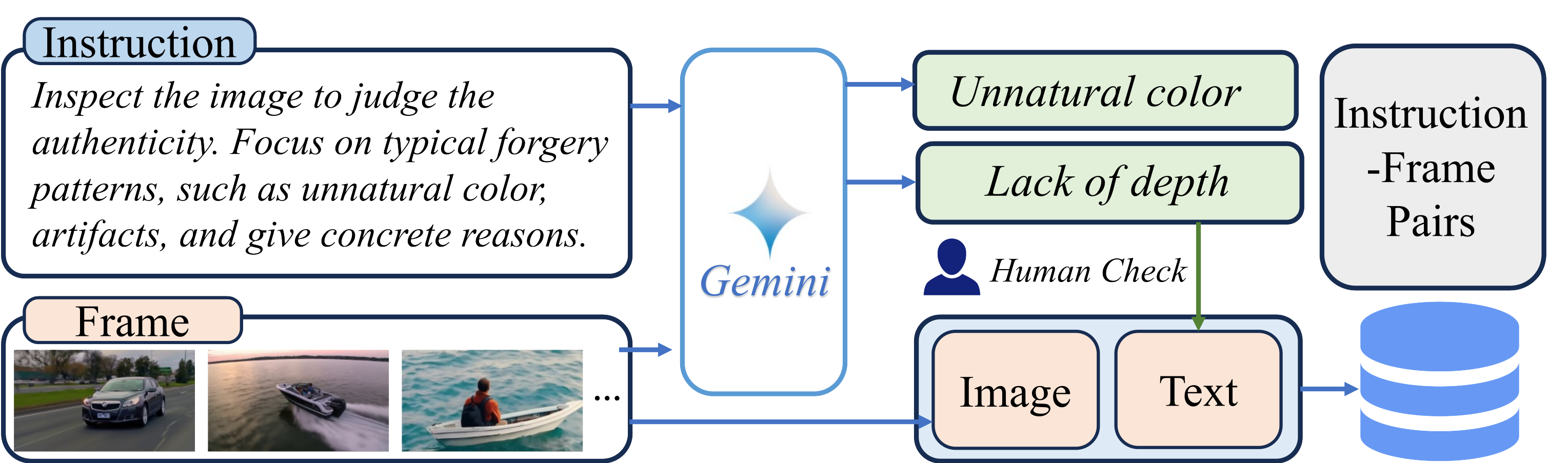}
    \caption{The pipeline of producing instruction-frame pairs. Gemini~\cite{team2023gemini} is leveraged to provide the forensics-oriented descriptions given the input video frames. After careful human inspection and adjustment, the collected pairs are utilized for instruction tuning of the MM branch.}
    \label{fig:dataset_rfrd}
\end{figure}

However, the selected training set does not support the finetuning of MLLM due to the lack of instruction-frame paired data. To alleviate the annotation burden, we resort to the powerful ability of Gemini~\cite{team2023gemini} to analyze the input video frames from the training set and provide detailed judgment and reasoning about their authenticity. This process achieves the automatic construction of a large number of instruction-frame pairs, enabling instruction-tuning in the MM branch at a friendly cost. Subsequently, we conduct a careful human inspection and adjustment to select high-quality instruction-frame pairs, comprising $3,479$ descriptions from $2,528$ real frames and $951$ fake frames, which are further augmented into $25$k conversations. Fig.~\ref{fig:dataset_rfrd} illustrates the pipeline of producing these instruction-frame pairs.

For a fair comparison between frame-level and video-level detectors, we adapt the frame-level baselines by employing two approaches to aggregate frame-level scores into video-level predictions: 1) all frame-level detection scores are averaged as a video-level prediction; 2) several sequential decoders, including LSTM~\cite{hochreiter1997long}, ViT~\cite{dosovitskiy2021image} and ViViT~\cite{arnab2021vivit}, are cascaded after frame-level detectors to capture temporal and spatial forgery traces in videos. Once trained on our dateset with the frame-level detectors kept frozen, these sequential models are capable of performing video-level prediction.


\subsection{Implementation Details}
\label{sec:expt_implementation_details}
Here we describe the detailed implementation of our MM-Det++. In ST branch,
we adopt Hybrid-ViT-B as the backbone, which consists of a ViT-B/$16$ on top of a ResNet-$50$, with the patch size of $14 \times 14$ and the hidden dimension of $M = 768$. We choose $N = 10$ as the length of the input frame sequence. 
The attention mechanism of FC-ViT is employed in all $12$ transformer blocks, with $10$ FC-tokens initialized from the pretrained class token of ViT~\cite{dosovitskiy2021image}. 
For positional embeddings, learnable spatial and temporal embeddings are separately trained and summed together. Specifically, all patches at the same location within frames share the same spatial embedding, while patches with the same timestep share the same temporal embedding. 
In MM branch, we utilize LLaVA v$1.5$~\cite{liu2024visual}, which consists of a CLIP~\cite{radford2021learning} encoder $\mathcal{E}_{V}$ (\textit{i.e.} CLIP-ViT-L-patch$14$-$336$) and a large language model $\mathcal{D}$ (\textit{i.e.} Vicuna-$7$b) for reasoning. We extract the hidden states from second-to-last layer from $\mathcal{E}_{V}$ as $\mathbf{H}_{v}$, where the sequence length $L = 577$ and the dimension $S = 1024$. The hidden state of LR-token from the last layer of $\mathcal{D}_{L}$ is extracted as $\mathbf{H}_{lr}^{\prime} \in \mathbb{R}^{D}$, where the dimension $D = 4096$.


For both training and inference, we use a single NVIDIA RTX $4090$ GPU. To finetune the MLLM with LoRA, we use an Adam optimizer with the learning rate $2e^{-5}$ for $10$ epochs, which takes approximately $40$ GPU hours. Afterward, we integrate LLaVA into MM-Det++ and train the entire framework in an end-to-end manner. $10\%$ of the training data is reserved as the validation set. 
For each video, $10$ consecutive frames are randomly sampled and cropped into $224 \times 224$ as input. The overall training is also performed with the Adam optimizer but using a learning rate of $1e^{-4}$ until convergence. The training time for this stage requires approximately $30$ GPU hours.

\setlength{\tabcolsep}{6pt} 

\renewcommand{\arraystretch}{1.1}
\begin{table*}[t]
  \caption{The video-level detection performance measured by AUC. The proposed MM-Det++ is compared with both frame-level and video-level detection methods. $^{*}$ means we append sequential decoders after frame-level detection methods for video-level detection. [Keys: \textcolor{red}{\textbf{Red}}: best result; \textcolor{blue}{Blue}: second best result].
  } 
  \label{tab:cross_dataset_evaluation_++}
  \centering
  \resizebox{\linewidth}{!}{
  \begin{tabular}{c|c|cccccccccc|c}
     \toprule
    \multicolumn{2}{|c|}{\multirow{2}{*}{\diagbox[width=3cm]{\textbf{Method}}{\textbf{Test Subset}}}} & \textbf{Video-} & \multirow{2}{*}{\textbf{SVD}} & \textbf{Zero-} & \multirow{2}{*}{\textbf{Sora}} & \multirow{2}{*}{\textbf{Pika}} & \textbf{Open-} & \multirow{2}{*}{\textbf{LDM}} & \textbf{Stable} & \textbf{Hun-} & \textbf{Cog-} & \multirow{2}{*}{\textbf{Average}}\\
    \multicolumn{2}{|c|}{} & \textbf{Crafter} & & \textbf{scope} & & & \textbf{Sora} & & \textbf{Video} & \textbf{Yuan} & \textbf{VideoX} &\\
    \midrule
    \multicolumn{13}{c}{\textbf{\textit{\normalsize Frame-level Detection Method$^{*}$}}} \\
    \midrule
    \multirow{4}{*}{CNN-Det}& - & $\color{gray}{\mathbf{99.9}}$ & $54.2$ & $71.5$ & $89.4$ & $93.6$  & $92.1$ & $80.1$ & $94.8$ & $55.1$ & $ 55.7$ & $78.6$\\
    & LSTM & $\color{gray}{\mathbf{99.9}}$ & $53.6$ & $73.5$ & $93.4$ & $91.7$  & $93.3$ & $\color{blue}{85.9}$ & $95.4$ & $54.2$ & $62.5$ & $80.3$\\
    & ViT & $\color{gray}{\mathbf{99.9}}$ & $62.5$ & $70.6$ & $94.9$ & $98.7$  & $94.2$ & $82.5$ & $97.6$ & $53.9$ & $59.8$ & $81.5$\\
    & ViViT & $\color{gray}{\mathbf{99.9}}$ & $55.1$ & $80.2$ & $88.5$ & $98.9$  & $93.3$ & $72.1$ & $97.9$ & $56.6$ & $59.0$ & $80.2$\\
    \midrule
    \multirow{4}{*}{Uni-Det}& - & $\color{gray}{\mathbf{99.9}}$ & $64.5$ & $79.8$ & $86.9$ & $94.8$ & $89.2$ & $77.9$ & $94.6$ & $87.6$ & $73.5$ & $84.9$\\
    & LSTM & $\color{gray}{\mathbf{99.9}}$ & $69.1$ & $80.8$ & $\color{blue}{95.0}$ & $99.0$ & $94.1$ & $60.1$ & $\color{blue}{98.2}$ & $89.6$ & $75.7$ & $86.2$\\
    & ViT & $\color{gray}{\mathbf{99.9}}$ & $60.7$ & $67.2$ & $79.3$ & $82.0$  & $82.3$ & $83.8$ & $94.0$ & $86.9$ & $75.0$ & $81.1$\\
    & ViViT & $\color{gray}{\mathbf{99.9}}$ & $71.3$ & $77.5$ & $86.9$ & $96.0$  & $85.4$ & $62.3$ & $96.0$ & $88.9$ & $72.4$ & $83.7$\\
    \midrule
    \multirow{4}{*}{HiFi-Net}& - & $\color{gray}{\mathbf{99.9}}$ & $73.6$ & $81.8$ & $86.2$ & $96.6$ & $88.9$  & $79.1$ & $95.5$ & $82.7$ & $78.6$ & $86.3$ \\
    & LSTM & $\color{gray}{\mathbf{99.9}}$ & $75.6$ & $81.1$ & $87.5$ & $95.8$ & $90.1$   & $82.6$ & $92.5$ & $85.5$ & $79.1$ & $87.0$\\
    & ViT & $\color{gray}{\mathbf{99.9}}$ & $\color{blue}{77.8}$& $82.9$ & $87.9$& $96.7$ & $92.4$  & $81.9$ & $96.8$ & $84.9$ & $80.3$ & $88.2$ \\
    & ViViT & $\color{gray}{\mathbf{99.9}}$ & $72.6$& $82.2$ & $85.4$& $91.8$ & $86.5$  & $81.2$ & $92.8$ & $81.8$ & $79.9$ & $85.4$\\
    \midrule
    \multirow{4}{*}{NPR}& - & $\color{gray}{\mathbf{99.9}}$ & $58.7$ & $70.1$ & $88.6$ & $99.1$ & $91.4$ & $84.1$ & $97.8$ & $65.8$ & $52.9$& $80.8$\\
    & LSTM & $\color{gray}{\mathbf{99.9}}$ & $59.6$ & $73.2$ & $94.3$ & $99.0$ & $\color{blue}{94.6}$ & $84.4$ & $\color{blue}{98.2}$ & $68.7$ & $51.8$ & $82.4$\\
    & ViT & $\color{gray}{\mathbf{99.9}}$ & $55.0$ & $65.8$ & $92.3$ & $98.8$ & $88.0$ & $83.7$ & $97.3$ & $62.0$ & $53.1$ & $79.7$\\
    & ViViT & $\color{gray}{\mathbf{99.9}}$ & $52.0$ & $70.8$ & $88.5$ & $97.4$  & $71.3$ & $84.1$ & $89.2$ & $65.8$ & $55.4$ & $77.4$\\
    \midrule
    \multirow{4}{*}{F3Net}& - & $\color{gray}{\mathbf{99.9}}$ & $62.1$ & $84.3$ & $75.8$ & $94.8$ & $79.4$ & $60.8$ & $90.0$ & $85.3$ & $75.9$& $80.8$\\
    & LSTM & $\color{gray}{\mathbf{99.9}}$ & $65.1$ & $81.7$ & $86.9$ & $96.2$ & $85.4$ & $72.8$ & $95.9$ & $87.7$ & $74.0$ & $84.6$\\
    & ViT & $\color{gray}{\mathbf{99.9}}$ & $68.2$ & $83.8$ & $85.4$ & $96.9$  & $84.3$ & $80.2$ & $96.2$ & $86.5$ & $73.8$ & $85.5$\\
    & ViViT & $\color{gray}{\mathbf{99.9}}$ & $65.4$ & $82.6$ & $88.8$ & $96.2$  & $86.0$ & $78.8$ & $94.1$ & $82.7$ & $70.8$ & $84.5$\\
    \midrule
    \multirow{4}{*}{DIRE}& - & $\color{gray}{\mathbf{99.9}}$ & $58.6$ & $59.8$ & $57.4$ & $68.9$ & $64.4$ & $57.1$ & $62.5$ & $59.8$ & $52.1$ & $64.1$\\
    & LSTM & $\color{gray}{\mathbf{99.9}}$ & $60.3$ & $52.6$ & $59.8$ & $77.3$ & $68.3$ & $53.0$ & $68.8$ & $60.2$ & $56.7$ & $65.7$\\
    & ViT & $\color{gray}{\mathbf{99.9}}$ & $59.5$ & $54.8$ & $63.0$ & $69.6$  & $59.5$ & $66.3$ & $64.5$ & $59.1$ & $58.2$ & $65.4$\\
    & ViViT & $\color{gray}{\mathbf{99.9}}$ & $58.2$ & $60.3$ & $63.0$ & $63.3$  & $61.3$ & $68.4$ & $60.5$ & $60.5$ & $59.2$ & $65.5$\\
    \midrule
    \multicolumn{13}{c}{\textbf{\textit{\normalsize Video-level Detection Method}}} \\
    \midrule
    \multicolumn{2}{c|}{TALL}  & $\color{gray}{\mathbf{99.9}}$ & $50.9$ & $54.9$ & $56.8$ & $57.0$  & $56.6$ & $73.0$ & $52.2$ & $54.2$ & $59.1$ &  $61.5$\\
    \multicolumn{2}{c|}{TS2-Net}  & $\color{gray}{\mathbf{99.9}}$ & $52.1$ & $51.8$& $55.7$ & $59.2$  & $55.9$ & $58.7$ & $59.1$& $52.4$ & $53.1$ & $59.8$\\
    \multicolumn{2}{c|}{MM-Det} & $\color{gray}{\mathbf{99.9}}$ & $75.1$ & $\color{blue}{86.0}$ & $92.9$ & $\color{blue}{99.1}$ & $86.0$  & $84.7$ & $96.0$ & $\color{blue}{90.4}$ & $\color{blue}{83.0}$ & $\color{blue}{89.3}$ \\
    \midrule
    
    \multicolumn{2}{c|}{MM-Det++ (Ours)} & $\color{gray}{\mathbf{99.9}}$ & $\color{red}{\mathbf{83.2}}$ & $\color{red}{\mathbf{86.7}}$ & $\color{red}{\mathbf{96.0}}$ & $\color{red}{\mathbf{99.3}}$ & $\color{red}{\mathbf{95.4}}$ & $\color{red}{\mathbf{86.2}}$ & $\color{red}{\mathbf{98.5}}$ & $\color{red}{\mathbf{96.1}}$ & $\color{red}{\mathbf{94.2}}$ & $\color{red}{\mathbf{93.5}}$\\
    \bottomrule
  \end{tabular}
  }
\end{table*}
\subsection{Experimental Results}
\label{sec:expt_main}

Tab.~\ref{tab:cross_dataset_evaluation_++} reports the video-level forgery detection performance, where our proposed MM-Det++ outperforms all other detectors, achieving an overall performance of $93.2\%$ on AUC. Note that the evaluation results on validation set of Video-Crafter is shown in this table, while the others are obtained on test subsets that remain unseen during training.

\textbf{First, we examine performance of frame-level detection methods.}
CNN-Det\cite{wang2020cnn} and Uni-Det~\cite{ojha2023towards}, which directly adopt widely used backbones (\textit{i.e.}, ResNet~\cite{wang2020cnn} and ViT~\cite{dosovitskiy2021image}), achieve the best AUCs of $81.5\%$ and $86.2\%$ on video-level forgery detection with sequential decoders built on ViT and LSTM, respectively. These results surpass the baseline of averaging frame-level scores, which confirm the necessity of appropriate sequential decoders for frame-level detection methods.
We further evaluate three advanced frame-level detectors that exploit the advantages of frequency information. HiFi-Net~\cite{guo2023hierarchical} employs a multi-branch feature extractor to capture artifacts across different domains, NPR~\cite{tan2024rethinking} explores correlations among neighboring pixels, and F3Net~\cite{qian2020thinking} extracts both global and local forgery traces with Discrete Cosine Transform (DCT). These methods achieve competitive performance on several test subsets. For instance, HiFi-Net ranks second on SVD with $77.8\%$, NPR achieves the best AUCs on Pika and Opensora with $99.6\%$ and $94.6\%$ respectively, while F3Net ranks second on Zeroscope with $83.8\%$. Although their performance demonstrates the effectiveness of frequency information for certain forgery traces, they are less effective at detecting the latest generative methods, such as Hunyuan ($87.7\%$) and CogVideoX ($75.9\%$), reflecting the limitations of frequency-based approaches in capturing subtle forgery cues.
In addition, we investigate a reconstruction-based method, DIRE~\cite{wang2023dire}, which employs a pretrained autoencoder of DDIM~\cite{song2021denoising} to detect diffusion traces by modeling frame-level reconstruction errors. However, even with a ViT as the sequential decoder, it achieves only $65.7\%$ AUC overall, validating that reconstruction-based methods are less effective for modern diffusion-generated videos. This limitation motivates us to remove the time consuming reconstruction process in MM-Det with the newly designed FC-ViT in MM-Det++.

In summary, although frame-level detection methods benefit from sequential decoders, the proposed MM-Det++ surpasses them in most cases, owing to its unified multimodal forgery learning that jointly detects diffusion traces in both spatial and temporal domains, and more importantly, leverages the powerful reasoning capabilities of MLLM.

\textbf{Then, we examine the performance of state-of-the-art video-level detectors.}
TALL~\cite{xu2023tall} relies on a tailored thumbnail layout to preserve spatial and temporal dependencies, but achieves only $61.5\%$ in AUC.
We attribute the limited performance to the thumbnail layout being specialized for face forgery videos, making it inadequate for handling general videos with diverse scenes.
TS2-Net~\cite{liu2022ts2} employs a token selection mechanism to capture informative spatio-temporal forgery traces, yet achieves only $59.8\%$ in AUC. We argue that its reliance on the selected local spatial semantics is insufficient for detecting subtle forgery traces.
Our precedent work MM-Det~\cite{song2024onlearning} achieves the second-best overall performance. However, the multi-round prediction and error reconstruction process compromise the detection efficiency, and the naive integration of multiple representations constrains its effectiveness in detecting diverse diffusion-generated videos. 
In comparison, our proposed MM-Det++ achieves the best overall performance, outperfomring the second-best method (\textit{i.e.} MM-Det) by $4.2\%$ in averaged AUC, with consistent improvements across different video generation approaches. This superiority stems from the two dedicated branches and the Unified Multimodal Learning module. The contribution of each component in forgery detection is detailed in our ablation study.

\subsection{Ablation Study}
\label{sec:expt_abla}

Tab.~\ref{tab:abla_study_mm++} demonstrates the ablation study that validates the impact of individual components in MM-Det++. 
Specifically, Model $1$ refers to CNN-Det with ViT appended as the sequential decoder, which achieves an average $87.7\%$ in AUC on our unseen test dataset. 
Model $2$ only utilizes the elaborately designed ST branch and achieves $99.9\%$ and $89.9\%$ on seen and unseen types of diffusion-generated videos, which justifies its effectiveness in aggregating global and local forgery information.
Model $3$ applies only the MM branch. Surprisingly, even with less competitive performance on seen forgery types ($99.7\%$), it surpasses the Model 2 on unseen forgery types ($90.8\%$ against $89.9\%$). This justifies the motivation for employing MLLMs in forensics, where their advanced reasoning capability contribute to improved generalization ability of the detector. Our full model (MM-Det++) integrates the spatio-temporal forgery representation from ST branch and multimodal forgery representation from MM branch via the Unified Multimodal Learning module. It achieves the best average AUC of $93.5\%$, surpassing the models with only ST branch or MM branch due to the merits of unified multimodal representation in forensics.

\setlength{\tabcolsep}{6pt}

\begin{table}[t]
  \caption{Ablation study of MM-Det++ in detection performance measured by AUC. \textit{Seen} and \textit{Unseen} denote the videos from seen and unseen generation methods in training. [Key: ST: Spatial-Temporal Branch; MM: Multimodal Branch; UML: Unified Multimodal Learning; \textbf{Bold}: Best.]}
  \label{tab:abla_study_mm++}
  \centering
    \resizebox{1\linewidth}{!}{
    \begin{tabular}{ccccccc}
    \toprule
    \multirow{2}{*}{\textbf{Models}} & \multirow{2}{*}{\shortstack[c]{\textbf{ST}}} & \multirow{2}{*}{\shortstack[c]{\textbf{MM}}} & \multirow{2}{*}{\textbf{UML}}  & \multirow{2}{*}{\textbf{Seen}} & \multirow{2}{*}{\textbf{Unseen}} & \multirow{2}{*}{\textbf{Average}}  \\
     & & & &   & \\
     \midrule

    $1$ & & & & $\mathbf{99.9}$& $86.0$& $87.7$\\[2pt]
    $2$ & \cmark & & & $\mathbf{99.9}$& $89.9$& $91.2$\\[2pt]
    $3$ &  & \cmark & &$99.7$&$90.8$&$91.9$\\[2pt]
    Ours & \cmark & \cmark  & \cmark &$\mathbf{99.9}$&$\mathbf{92.6}$&$\mathbf{93.5}$\\
    \bottomrule
  \end{tabular}}
\end{table}

\setlength{\tabcolsep}{8pt}
\begin{table}[tb]
  \caption{Detection performance of using different large language models in MM-Det++ [Keys: \textbf{Bold}: Best.].}
  \label{tab:abla_study_llms}
  \centering
    \resizebox{1\linewidth}{!}{

  \begin{tabular}{{c}{c}ccc}
    \toprule
    \multirow{2}{*}{\textbf{Models}} & \multirow{2}{*}{\textbf{LLMs}}  & \multirow{2}{*}{\textbf{Seen}}  & \multirow{2}{*}{\textbf{Unseen}}  & \multirow{2}{*}{\textbf{Average}}  \\
    & & & & \\
    \midrule
    4 & Mistral-$7$B  & $\mathbf{99.9}$ & $91.4$ & $92.2$ \\[2pt]
    5 & Vicuna-$13$B & $\mathbf{99.9}$ & $\mathbf{92.8}$ & $\mathbf{93.7}$ \\[2pt]
    Ours &  Vicuna-$7$B & $\mathbf{99.9}$ & $92.6$ & $93.5$ \\


  \bottomrule
  \end{tabular}
  }
\end{table}






Moreover, we modify the LLM backbone in LLaVA with alternatives and report their performance in Tab.~\ref{tab:abla_study_llms}.
For a fair comparison, when employing different LLMs, we keep all other components (\textit{i.e.} the ST branch, MM branch, and UML module) identical to the base model. 
Specifically, Model $4$ with Mistral-7B~\cite{jiang2023mistral} and our MM-Det++ with Vicuna-7B~\cite{zheng2023judging} have the similar number of parameters, yet it yields an average AUC lower by $1.3\%$. This observed performance gap may be attributed to structural differences, where Vicuna-7B employs a casual attention mechanism that aligns well with the learnable reasoning process. In comparison, Mistral-7B adopts a sliding window attention that restricts attention to tokens within a fixed window for next-token prediction, which constrains the token capacity to exploit subtle forgery traces.
It is worth noting that with a larger LLM (\textit{e.g.} Vicuna-$13$B~\cite{zheng2023judging}), Model $5$ outperforms our MM-Det++ by $0.2\%$ in average AUC. This finding substantiates the general effectiveness of exploiting the reasoning capabilities of LLMs for forgery detection, and validates that the scaling law also holds for forgery detection, as larger MLLMs tend to exhibit stronger reasoning capabilities. We believe this ablation study clearly demonstrates the superiority of leveraging MLLMs for reasoning forgery traces in forensics, and also shows that the proposed MM-Det++ can be  extended with larger MLLMs to further improve its detection performance. Considering the trade-off between effectiveness and efficiency in forgery detection, we adopt the Vicuna-7B in LLaVA as the default configuration in MM-Det++.

\begin{figure*}[t]
    \centering
    \captionsetup[subfloat]{labelfont=footnotesize,textfont=footnotesize}
    \renewcommand{\thesubfigure}{{a}}
    \subfloat[\footnotesize Clustering visualization of spatio-temporal forgery representations]{\includegraphics[width=0.325\linewidth]{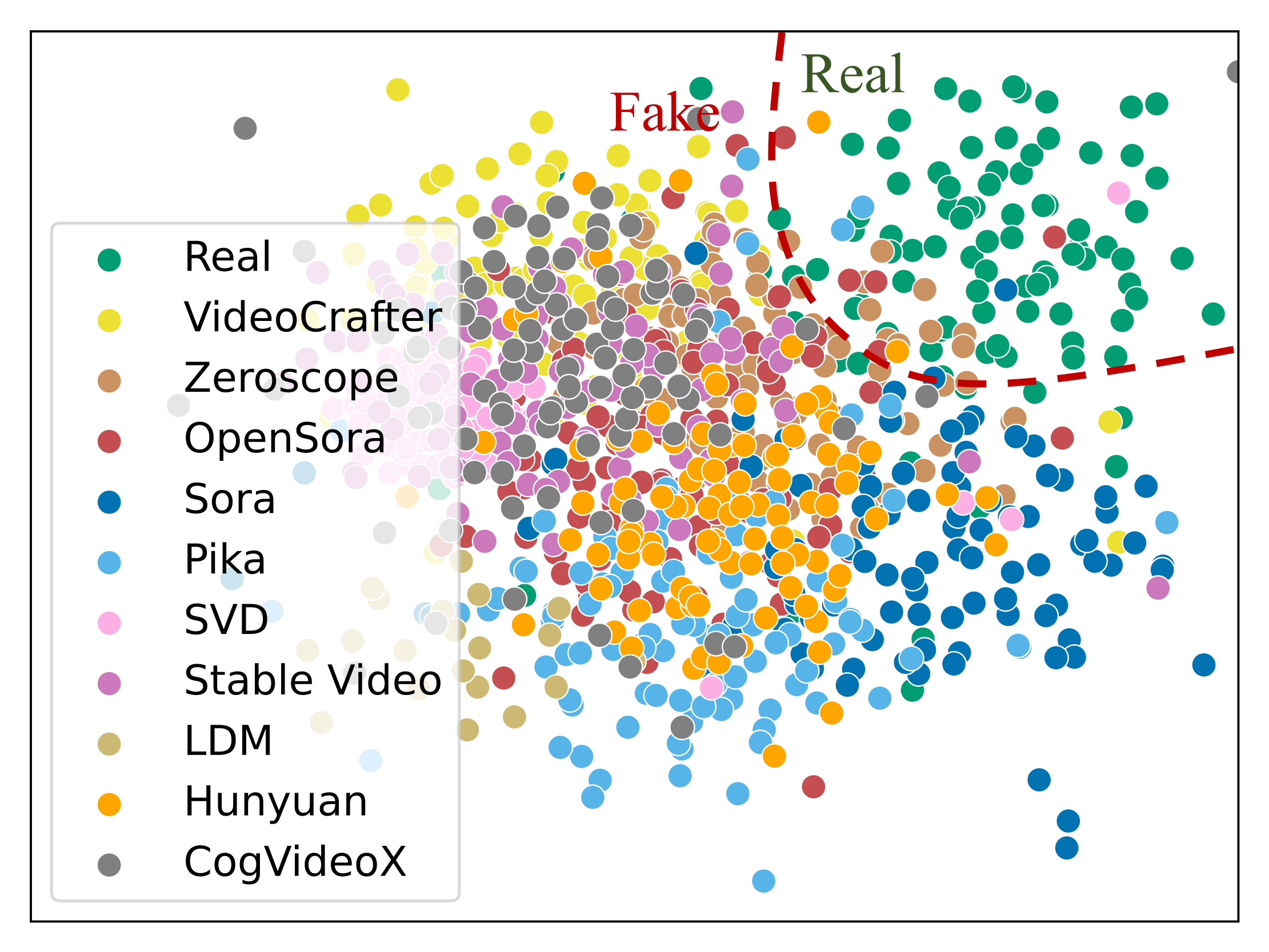}}
    \hfill
    \renewcommand{\thesubfigure}{{b}}
    \subfloat[\footnotesize Clustering visualization of multimodal forgery representations]{\includegraphics[width=0.325\linewidth]{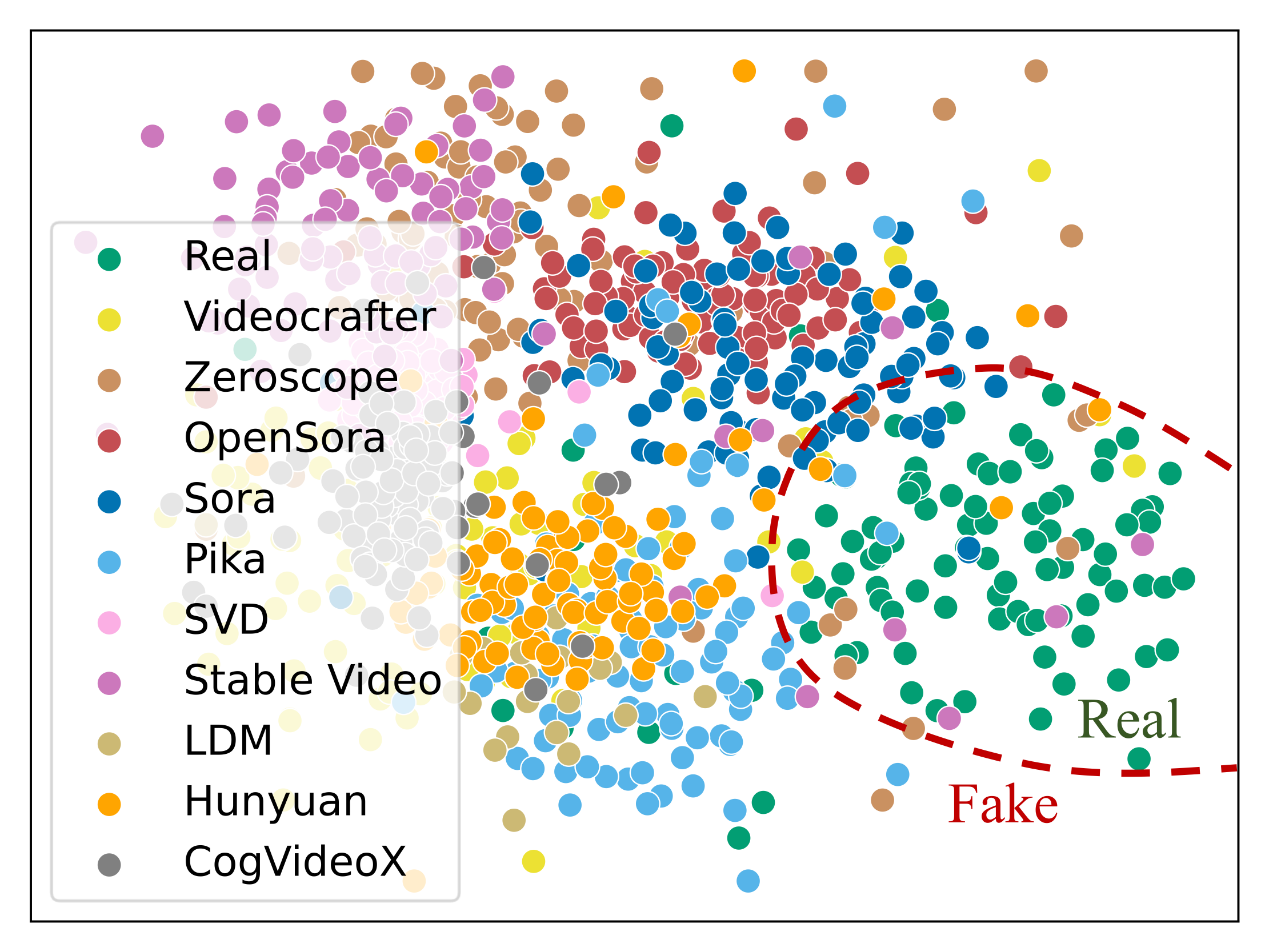}}
    \renewcommand{\thesubfigure}{{c}}
    \subfloat[\footnotesize Detection accuracy of reasoning representations from different layers in LLaVA.]{\includegraphics[width=0.315\linewidth]{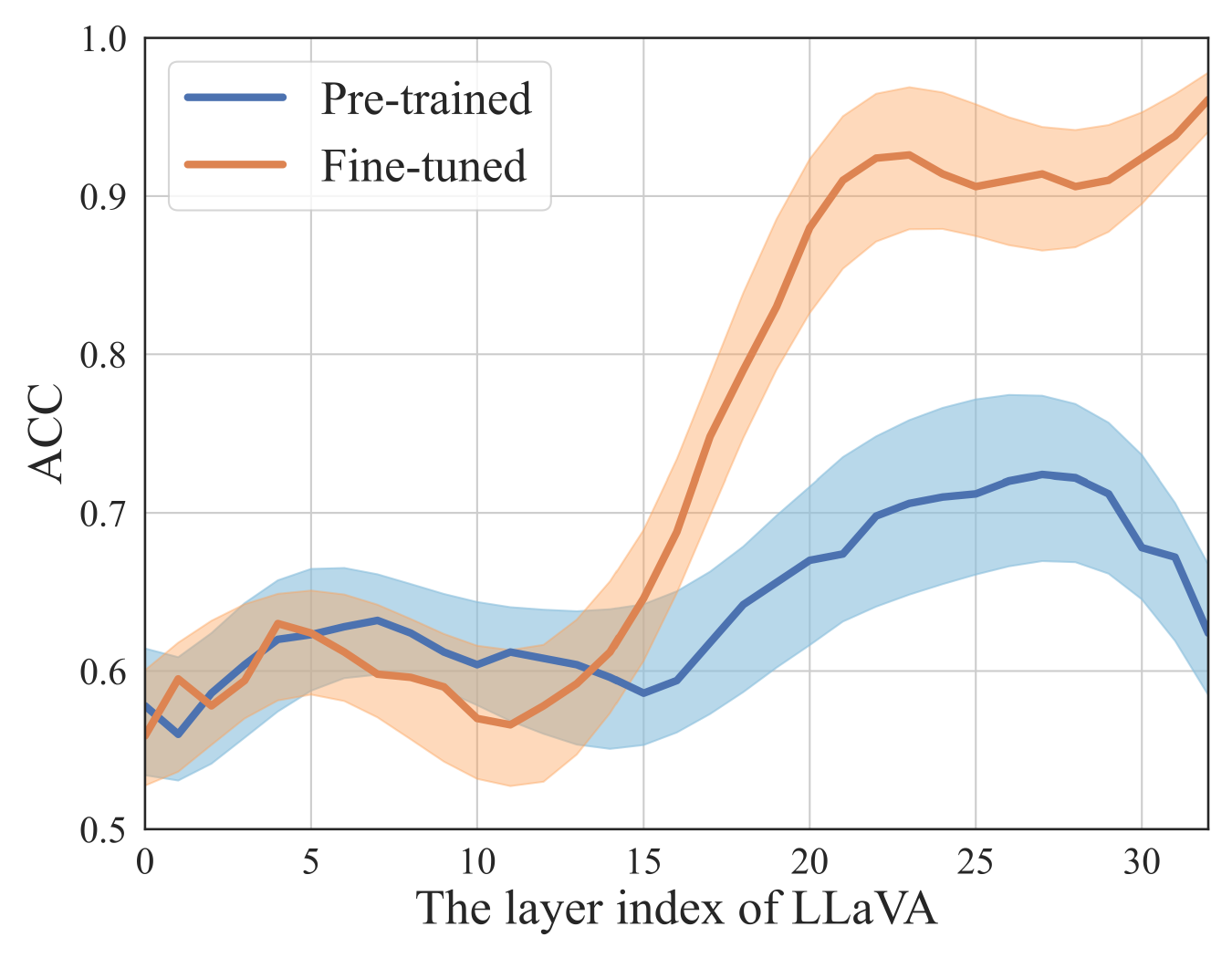}}
    \caption{t-SNE~\cite{van2008visualizing} (a)(b) Visualizations of spatio-temporal and multimodal forgery representations respectively, both showing clear boundaries between real and fake videos. (c) Evaluation of the optimal reasoning representation in LLaVA.}
    \label{fig:expt_clustering}
\end{figure*}

\subsection{Clustering Visualization of Forgery Representations}
\label{sec:expt_mmfr_analysis}
Here we visualize the obtained spatio-temporal forgery representation from ST branch and multimodal forgery representation from MM branch via t-SNE~\cite{van2008visualizing} to validate their effectiveness in detecting diffusion-generated videos. To achieve this visualization, we sample 100 videos from each test subset. Fig.~\ref{fig:expt_clustering}\textcolor{blue}{a} and Fig.~\ref{fig:expt_clustering}\textcolor{blue}{b} demonstrate the clustering visualizations of these two forgery representations respectively, where clear boundaries enable effective separation between real and fake videos. We observe that several clusters overlap with each other, which may be attributed to the structural similarity of related video generation methods. In addition, the overlaid clusters of spatio-temporal forgery representations differ from those of multimodal forgery representations. The underlying reason may lie in the distinct perspectives of two representations, with one capturing forgery traces from spatial and temporal domains of a video and the other leveraging the multimodal reasoning capabilities of MLLMs to handle forgery cases that necessitate semantic analysis. This observation also justifies the efficacy of unifying these two representations.

\begin{figure}[t]
    \centering
    \includegraphics[width=0.96\linewidth]{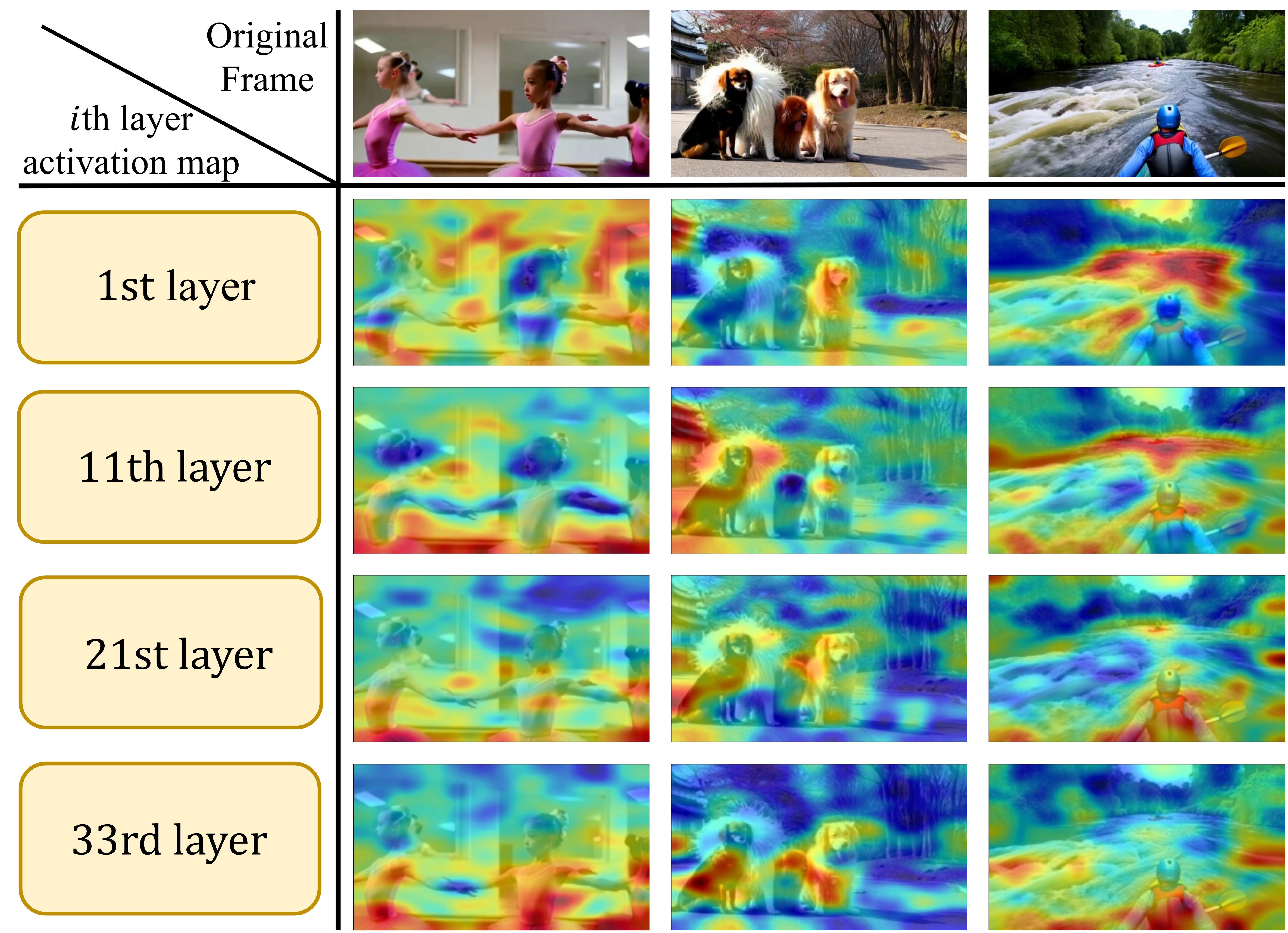}
    \caption{Activation maps from different layers in LLaVA, calculated by cosine similarity between the reasoning representation and the patch-level cross-modal representation.}
    \label{fig:vis_llava_mmfr_act}
\end{figure}

\subsection{Analysis of Reasoning Representation} 
\label{sec:expt:mllm_reasoning}
To find the optimal reasoning representation, we quantify the detection ability of the output from each LLM layer in LLaVA. To achieve this, we first randomly sample $1,000$ key frames from fake videos generated by SVD~\cite{rombach2022high}, and $1,000$ key frames from real videos that share the same semantics. The LLM (\textit{i.e.}, Vicuna-7B~\cite{touvron2023llama}) in LLaVA has 33 layers in total. For comparison, we extract the representations from the $i$th layer as $\mathbf{f}^{o}_i$, $i \in [1, 33]$ in both pre-trained and fine-tuned LLaVA by our DVF dataset, where the k-Means clustering algorithm is employed on $\mathbf{f}^{o}_i$. Fig.~\ref{fig:expt_clustering}\textcolor{blue}{c} demonstrates the detection accuracy of each layers in LLaVA. It can be seen that reasoning representations after $22$nd layers from fine-tuned LLaVA exhibit detection accuracies above $90\%$, surpassing the pre-trained counterpart by $15\%$. This phenomenon provides an important insight that the latter layers of MLLMs are capable of representing forgery information by reasoning over video semantics. The accuracy gap between the pre-trained and fine-tuned LLaVA also highlights the significance of instruction tuning for downstream tasks.

\setlength{\tabcolsep}{2pt}
\begin{table}[t]
  \caption{Detection performance of detectors with different representations in the Unified Multimodal Learning module measured by AUC. [Keys: Contra.: Contrastive Loss; CMR: Cross-Modal Representation; UMR: Unified Multimodal Representation; \textbf{Bold}: Best.].}
  \label{tab:abla_study_mm_adapter}
  \centering
    \resizebox{1\linewidth}{!}{
    \begin{tabular}{cccccccc}
    \toprule
    \multirow{2}{*}{\textbf{Models}} & 
    \multicolumn{3}{c}{\textbf{Representations}} & \textbf{Strategy} & \multirow{2}{*}{\textbf{Seen}} & \multirow{2}{*}{\textbf{Unseen}} & \multirow{2}{*}{\textbf{Average}}  \\
    \cmidrule{2-4} \cmidrule{5-5}
     & CLIP & CMR  & UMR & Contra. & & & \\[2pt]
    \midrule
    $6$ & \cmark & & &  & $\mathbf{99.9}$& $86.3$& $88.0$\\[2pt]
    $7$ &  & \cmark &  &  &$\mathbf{99.9}$&$91.1$&$92.2$\\[2pt]
    $8$ &  & & \cmark  & &$\mathbf{99.9}$&$91.9$&$92.9$\\[2pt]
    Ours &  &   & \cmark &\cmark & $\mathbf{99.9}$&$\mathbf{92.6}$&$\mathbf{93.5}$ \\
    \bottomrule
  \end{tabular}}
\end{table}

In addition, we select reasoning representations from $1${st}, $11${th}, $21${st}, and $33${rd} (the last) layers, and present their activation maps computed via cosine similarity between the reasoning representation $\mathbf{H}_{lr}^{\prime}$ and the patch-level cross-modal representation $\mathbf{H}_{c}^{p}$ in Fig.~\ref{fig:vis_llava_mmfr_act}. Specifically, the activation maps from the former layers (\textit{i.e.} $1${st} and $11${th}) cannot identify the Regions of Interest (ROIs), indicating that representations from these layers are less effective in reasoning forgery traces. 
On the contrary, representations from latter layers (\textit{i.e.} $21${st}, and $32${nd}) progressively pay attention to forgery ROIs such as the unreasonable appearance of humans and animals. This visualization further validates the advantage and importance of incorporating reasoning representations from the latter layers of MLLMs for video forgery detection. Our MM-Det++ utilizes the output from the last layer of LLaVA as the reasoning representations by default.

\begin{figure}[t]
    \flushright
    \begin{overpic}[width=0.94\linewidth]{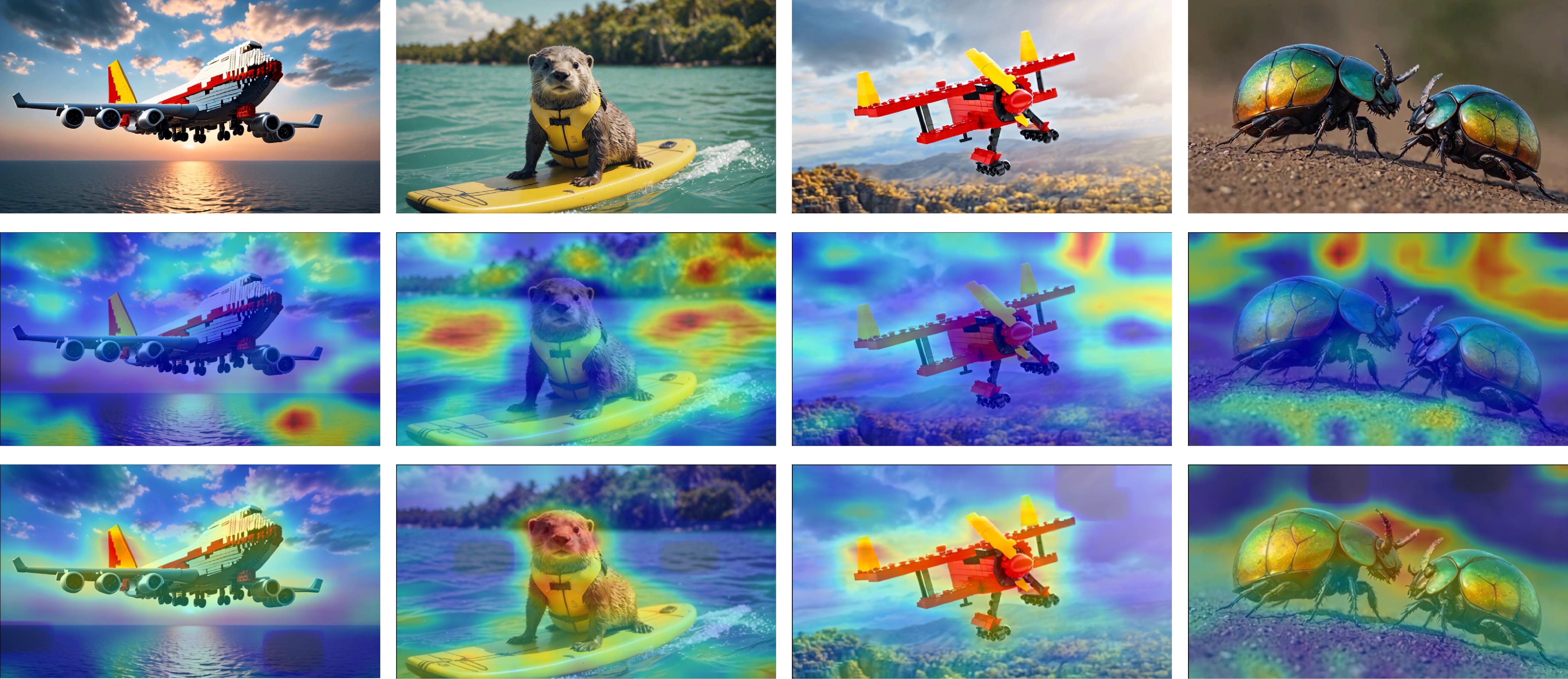}
    \put (-3.5, 34) {\footnotesize \begin{turn}{90}Ori.\end{turn}}
    \put (-3.5, 17.5) {\footnotesize \begin{turn}{90}CLIP\end{turn}}
    \put (-3.5, 3) {\footnotesize \begin{turn}{90}CMR\end{turn}}

    \vspace{2mm}
    \end{overpic}
    \caption{Visualization of the visual representation from a pre-trained CLIP encoder~\cite{radford2021learning} and the cross-modal representation in the Unified Multimodal Learning module. The activation maps are demonstrated based on the magnitude in each patch. [Key: Ori: Original, CMR: Cross-Modal Representation.]}
    \label{fig:expt_mm_adapter_act}
\end{figure}

\subsection{Analysis of Unified Multimodal Learning}
\label{sec:expt_lr_umlearning}

We analyze the importance of Unified Multimodal Learning via performance comparisons between different visual and reasoning representations.
Tab.~\ref{tab:abla_study_mm_adapter} demonstrates the detection performance of the visual representation $\mathbf{H}_{v}$ from the pre-trained CLIP encoder, the cross-modal representation $\mathbf{H}_{c}$ and the unified multimodal forgery representation $\mathbf{H}_{um}$.
Model $6$ only uses the CLIP visual representation. Although it performs well on seen forgery types, it falls short on unseen ones ($86.3\%$ in AUC), revealing its limited generalization ability. By effectively integrating the CLIP visual representation with the MLLM reasoning representation, the cross-modal representation allows Model $7$ to attain better performance on unseen forgery types ($91.1\%$ in AUC). Equipped with both spatio-temporal forgery representation and multimodal forgery representation, the unified forgery representation enables Model $8$ to achieve a $+0.8\%$ improvement on unseen forgery types. In addition, we evaluate the contribution of contrastive learning, which yields a further $+0.6\%$ improvement in AUC. This justifies that aligning visual and reasoning spaces benefits forgery representations in distinguishing fake from real videos.


Furthermore, Fig.~\ref{fig:expt_mm_adapter_act} visualizes activation maps of CLIP visual representations and cross-modal representations, where the shown fake videos are selected from the StableVideo subset in DVF. Since UMR is not a patch-level representation, we are unable to visualize its activation maps. Notably, the activation maps from CLIP visual representations seem noisy without any concrete semantic indication (sparsely focusing on the background region), whereas those from cross-modal representations highlight the potential objects with forgery traces (\textit{e.g.}, irregular and over-saturated planes, unrealistic dog and insects). By incorporating semantic reasoning and analysis, the cross-modal representation exhibits the strong capability in detecting diffusion-generated videos.

\begin{figure}[t]
    \flushright
    \begin{overpic}[width=0.95\linewidth]{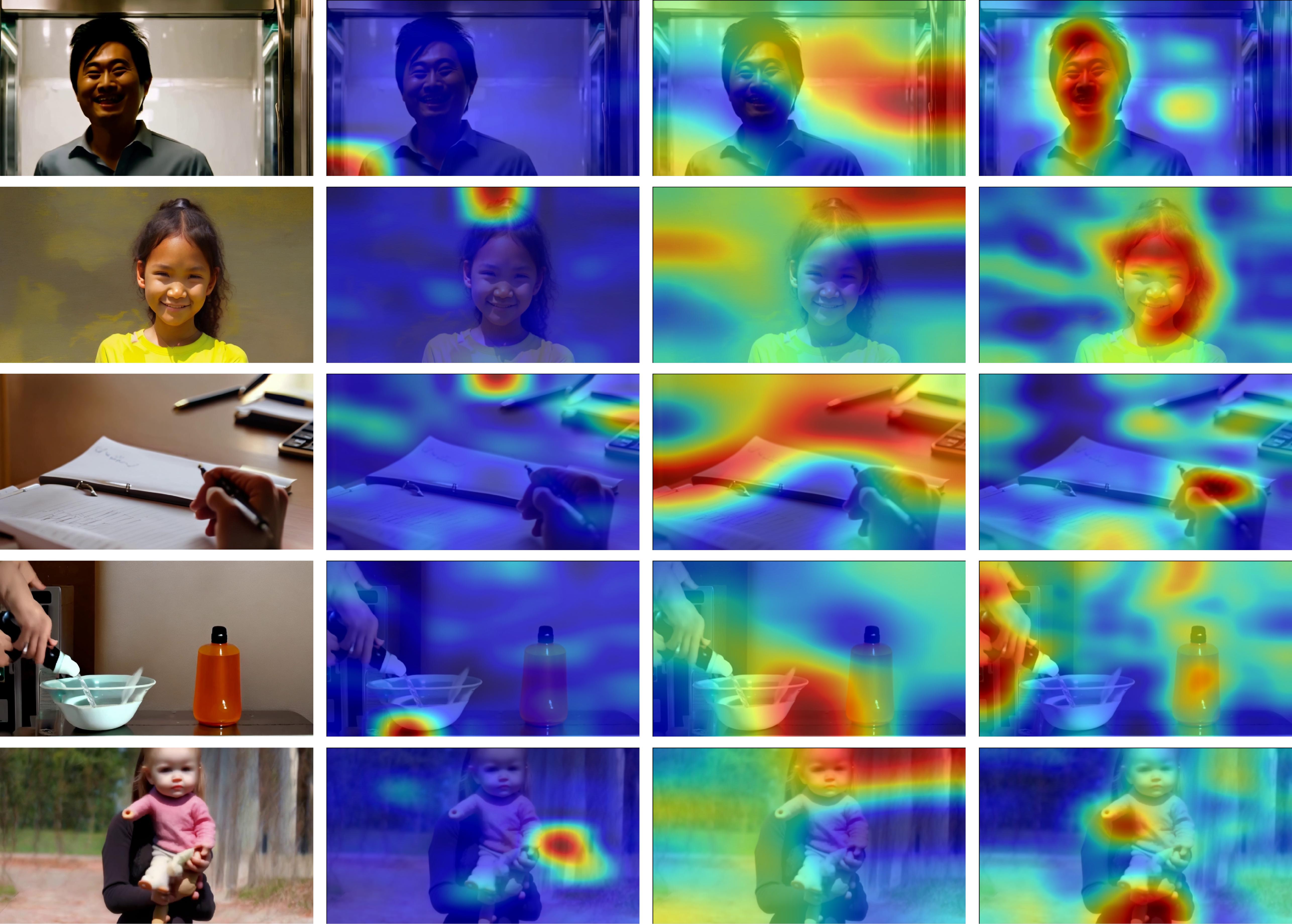}
        \put (-3.5, 59) {\footnotesize \begin{turn}{90}Frame 1\end{turn}}
        \put (-3.5, 44.5) {\footnotesize \begin{turn}{90}Frame 2\end{turn}}
        \put (-3.5, 30) {\footnotesize \begin{turn}{90}Frame 3\end{turn}}
        \put (-3.5, 15.5) {\footnotesize \begin{turn}{90}Frame 4\end{turn}}
        \put (-3.5, 1) {\footnotesize \begin{turn}{90}Frame 5\end{turn}}
        \put (7.5,-3.2) {\footnotesize Original}
        \put (35,-3.2) {\footnotesize ViT}
        \put (59,-3.2) {\footnotesize ViViT}
        \put (85,-3.2) {\footnotesize Ours}

    \end{overpic}
    \vspace{3mm}
    \caption{Visualization of spatial forgery activations from  ViT~\cite{dosovitskiy2021image}, ViViT~\cite{arnab2021vivit}, and our FC-ViT. Key frames from each video are selected.}
    \label{fig:expt_st_s}
\end{figure}

\begin{figure}[t]
    \flushright
    \begin{overpic}[width=0.94\linewidth]{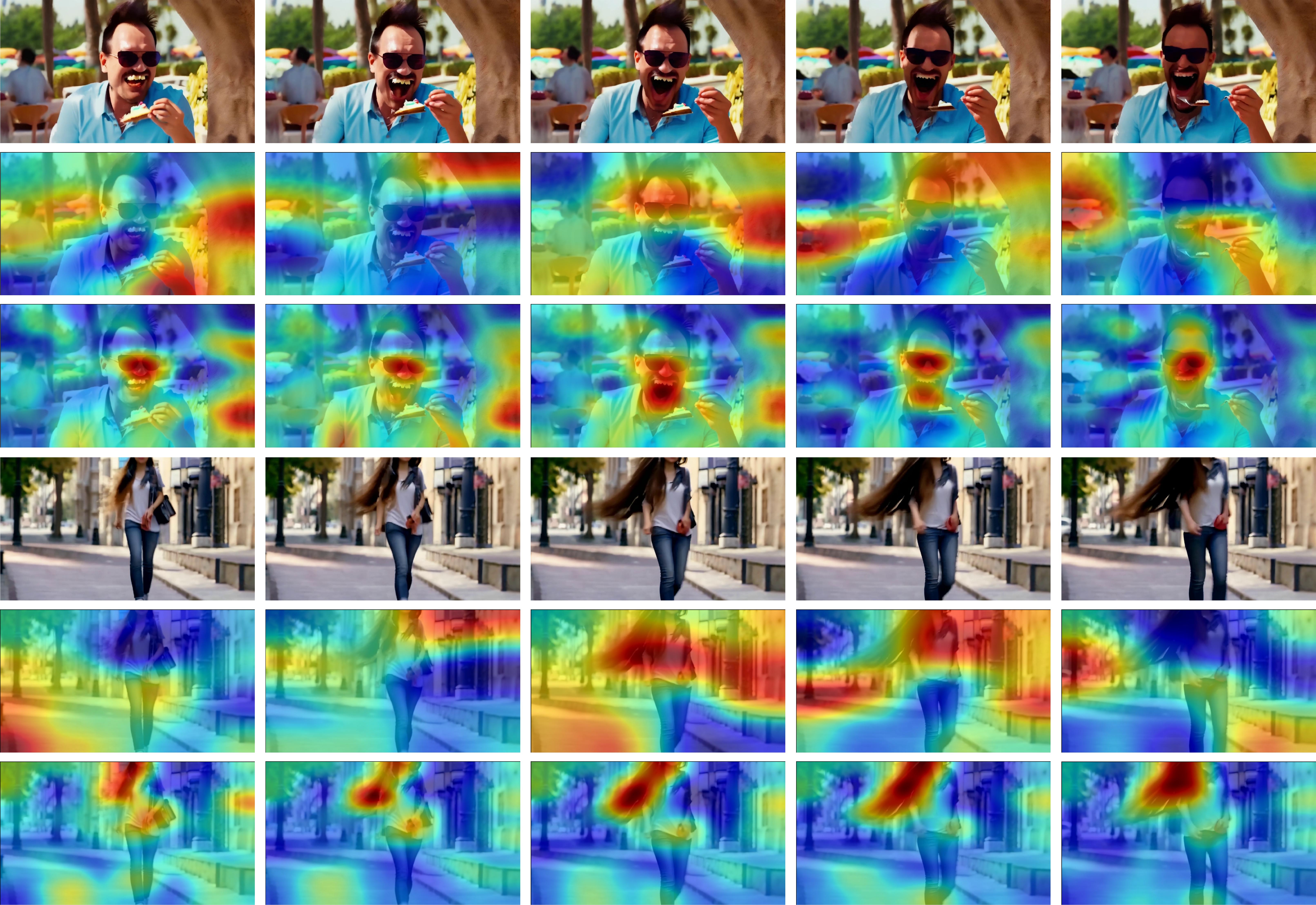}
        \put (-3.5, 58) {\footnotesize \begin{turn}{90}Video 1\end{turn}}
        \put (-3.5, 48) {\footnotesize \begin{turn}{90}ViViT\end{turn}}
        \put (-3.5, 36.5) {\footnotesize \begin{turn}{90}Ours\end{turn}}
        \put (-3.5, 23) {\footnotesize \begin{turn}{90}Video 2\end{turn}}
        \put (-3.5, 12.5) {\footnotesize \begin{turn}{90}ViViT\end{turn}}
        \put (-3.5, 1.5) {\footnotesize \begin{turn}{90}Ours\end{turn}}
        \put (5,69.5) {\footnotesize Time 1}
        \put (25.2,69.5) {\footnotesize Time 3}
        \put (45.4,69.5) {\footnotesize Time 5}
        \put (65.6,69.5) {\footnotesize Time 7}
        \put (85.8,69.5) {\footnotesize Time 9}
    \end{overpic}
    \vspace{2mm}
    \caption{Visualization of temporal forgery activations from ViViT~\cite{arnab2021vivit} and our FC-ViT. The time step is set to 2.
    }
    \label{fig:expt_st_t}

\end{figure}
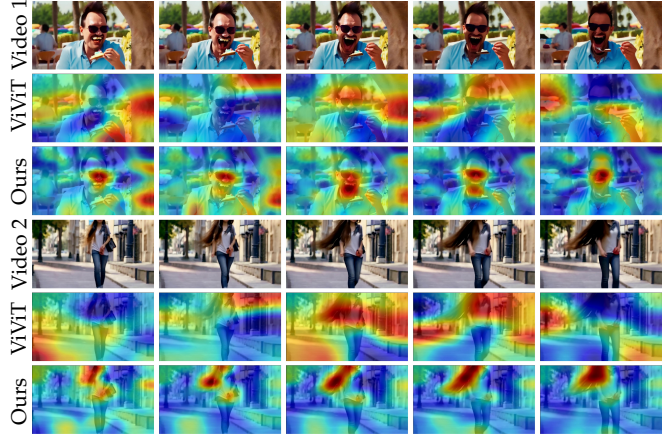

\subsection{Analysis of ST Branch}
\label{sec:expt_st_t}

To validate the effectiveness of spatio-temporal forgery representations from our ST branch and other representative spatiotemporal baselines such as ViT~\cite{dosovitskiy2021image} and ViViT~\cite{arnab2021vivit}, 
following Pang \textit{et al.}~\cite{pang2024frozen}, we visualize the activation maps from the last layer of the compared methods according to the magnitude with L2-norm. Fake videos are selected from VideoCrafter~\cite{chen2023videocrafter1}.

In Fig.~\ref{fig:expt_st_s}, we showcase the activated regions from a spatial perspective. Specifically, ViT~\cite{dosovitskiy2021image} and ViViT~\cite{arnab2021vivit} activate ambiguous regions that do not align with any evident spatial artifacts, manifesting their limited ability to capture spital forgery traces. For example, unnatural face shapes (Frame $1$ and $2$) and distorted body shapes (Frame $3$, $4$ and $5$) are not activated by them. In contrast, our FC-VIT shows precise activation on these forgery ROIs, mainly due to the proposed FC-tokens that capture forgery traces by aggregating spatio-temporal information in videos. In addition, Fig.~\ref{fig:expt_st_t} demonstrates the activated regions from a temporal perspective, where our  FC-ViT is compared with ViViT~\cite{arnab2021vivit} to evaluate the representation ability of capturing temporal forgery traces. ViViT fails to follow the forgery ROIs in different timestamps, for example facial artifacts in Video $1$ and unnatural hairs in Video $2$, while FC-VIT exhibits the promising capability in tracking forgery traces over time.

The thorough visualizations not only validate the superiority of our FC-VIT over the compared methods, but also disclose the effectiveness of FC-token design, which aggregates the spatio-temporal information in videos to reliably represent forgery traces.

\setlength{\tabcolsep}{4pt}
\begin{table}[t]
	\caption{Analysis of robustness on representative post-processing attacks measured by AUC. [Key: \textbf{Bold}: Best.]}
	
	\centering
	\resizebox{1.\linewidth}{!}{
		
		\begin{tabular}{ccccccc}
			\toprule
			\multirow{3}{*}{\textbf{Method}} & \multicolumn{6}{c}{\normalsize \textbf{Post-processing Attacks}} \\
			\cmidrule{2-7}
			& \multirow{2}{*}{\textbf{N/A}} & \textbf{Blur} & \textbf{JPEG} & \textbf{Resizing} & \textbf{Rotation} & \multirow{2}{*}{\textbf{Mixed}} \\
			& & $\sigma=3$ & $Q=70$ &  $0.7\times$ & $90^\circ$ & \\
			\midrule
			F3Net & $78.9$ & $70.1$ & $74.8$ & $69.2$ & $75.1$ & $72.0$ \\[2pt]
			HiFi-Net & $87.6$ & $78.7$ & $84.3$ & $80.2$ & $81.6$ & $80.9$ \\[2pt]
			Uni-Det & $88.4$ & $78.6$ & $82.1$ & $78.1$ & $81.7$ & $79.6$ \\[2pt]
			MM-Det & $88.9$ & $81.1$ & $83.6$ & $83.8$ & $84.7$ & $83.1$ \\[2pt]
			MM-Det++ & $\mathbf{95.3}$ & $\mathbf{89.8}$ & $\mathbf{93.4}$ & $\mathbf{91.9}$ & $\mathbf{93.2}$ & $\mathbf{91.8}$ \\
			\bottomrule
		\end{tabular}
	}
	\label{tab:expt_robustness}
\end{table}

\subsection{Analysis of Robustness}
\label{sec:expt_robustness}
To validate the robustness of our MM-Det++, we utilize the representative post-processing operations to imitate the potential attacks in real scenarios, including Gaussian blur ($\sigma = 3$), JPEG compression (quality $Q = 70$), resizing ($0.7\times$), rotation ($90^\circ$), and a mixture of all attacks (randomly select one type of attacks for each video). The test data consists of 500 real videos from Internvid-10M~\cite{wang2024internvid} and 500 fake videos from OpenSora in our DVF. As shown in Tab.~\ref{tab:expt_robustness}, both the frequency-based detectors (F3Net~\cite{qian2020thinking} and HiFi-Net~\cite{guo2023hierarchical}) and CLIP-based detector (Uni-Det~\cite{ojha2023towards}) exhibit significant performance degradation under these postprocessing attacks. The most severe drops occur under resizing, with F3Net decreasing by $9.7\%$ and Uni-Det by $10.3\%$, highlighting their vulnerability to such attack.

In contrast, our previous work, MM-Det~\cite{song2024onlearning}, achieves the second-best AUC of $83.1\%$ on the mixture of all attacks, while the  improved MM-Det++ attains the best AUC of $91.8\%$, with only a $3.5\%$ performance drop compared to the test without any attacks. We attribute this robustness primarily to the powerful and  comprehensive reasoning capabilities of MLLMs, which analyze video semantics to identify forgery traces. More importantly, this ability inherently complements our ST branch, which focuses on detecting forgery traces associated with relatively low-level cues.



\section{Conclusion}
In this paper, we introduce MM-Det++, a novel multimodal dual-branch detector for diffusion-generated video detection. The ST branch employs the proposed FC-ViT with FC tokens to capture spatio-temporal cues, while the MM branch harnesses the advanced comprehension and reasoning capabilities of MLLMs to acquire the multimodal forgery representation from a semantic perspective. In this end, a UML module further consolidates these complementary representations into a coherent feature space, enhancing the overall generalization ability of MM-Det++. To facilitate broader research in video forensics, we elaborately construct the DVF dataset, a comprehensive semantic-independent benchmark designed to support the detection of diffusion-generated videos produced by diverse methods. Extensive experiments demonstrate the superiority of MM-Det++ over existing detectors and highlight the effectiveness of our unified multimodal forgery learning.

\noindent\textbf{Limitations}: Despite its remarkable performance, we acknowledge that MM-Det++ may incur significant computational overhead and resource consumption, primarily due to its reliance on powerful MLLMs as the backbone. However, we remain optimistic that this challenge can be alleviated with the development of more efficient and lightweight MLLMs, as well as advanced techniques in model compression and distillation.

{
\small
\bibliographystyle{IEEEtran}
\bibliography{main}
}

\begin{IEEEbiography}[{\includegraphics[width=1in,height=1.25in,clip,keepaspectratio]{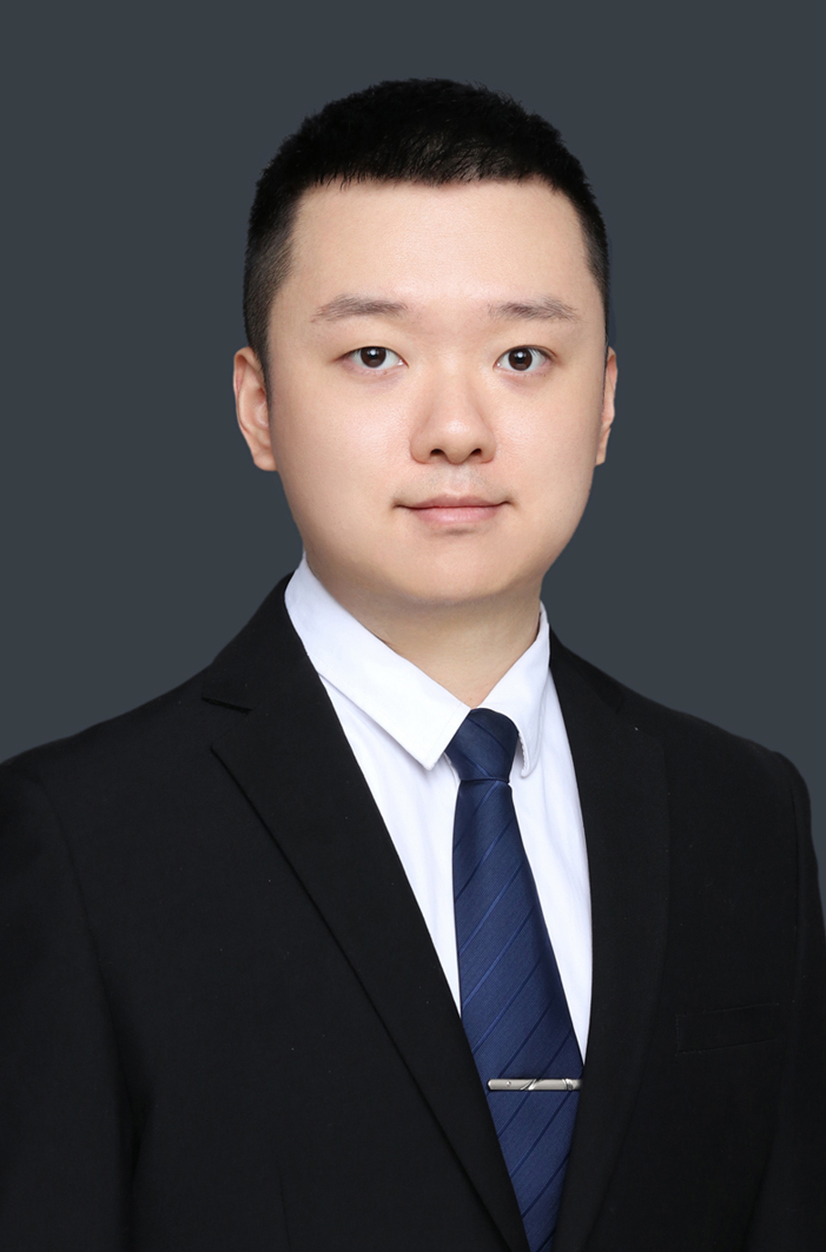}}]{Xiaohong Liu} (Member, IEEE) received the Ph.D. degree in electrical and computer engineering from McMaster University, Canada, in 2021. He is currently an associate professor with the School of Computer Science, Shanghai Jiao Tong University, China. His research interests lie in computer vision and multimedia. He has published over 100 academic papers in top journals and conferences, and has been recognized among the Stanford Top 2\% Scientists (2025), the Microsoft Research Asia StarTrack Scholars (2024), Chinese Government Award for Outstanding Self-financed Students Abroad (2021), and Borealis AI Fellowships (2020). He serves as an associate editor of the ACM TOMM.
\end{IEEEbiography}

\begin{IEEEbiography}[{\includegraphics[width=1in,height=1.25in,clip,keepaspectratio]{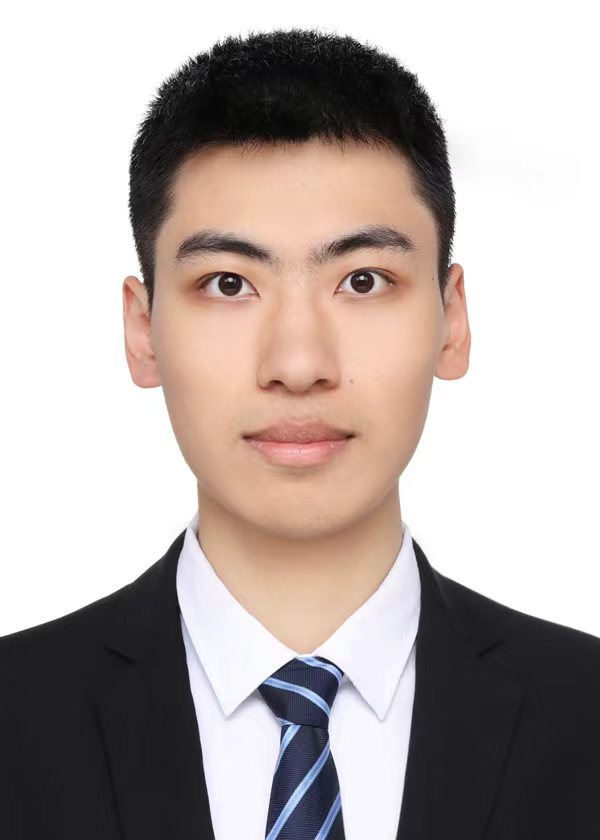}}]{Xiufeng Song} received the BS degree in computer science and technology at Shanghai Jiao Tong University, Shanghai, China, in 2023. He is currently working toward the master’s degree in multimodal forensics with the School of Computer Science, Shanghai Jiao Tong University, Shanghai, China.
\end{IEEEbiography}


\begin{IEEEbiography}[{\includegraphics[width=1in,height=1.25in,clip,keepaspectratio]{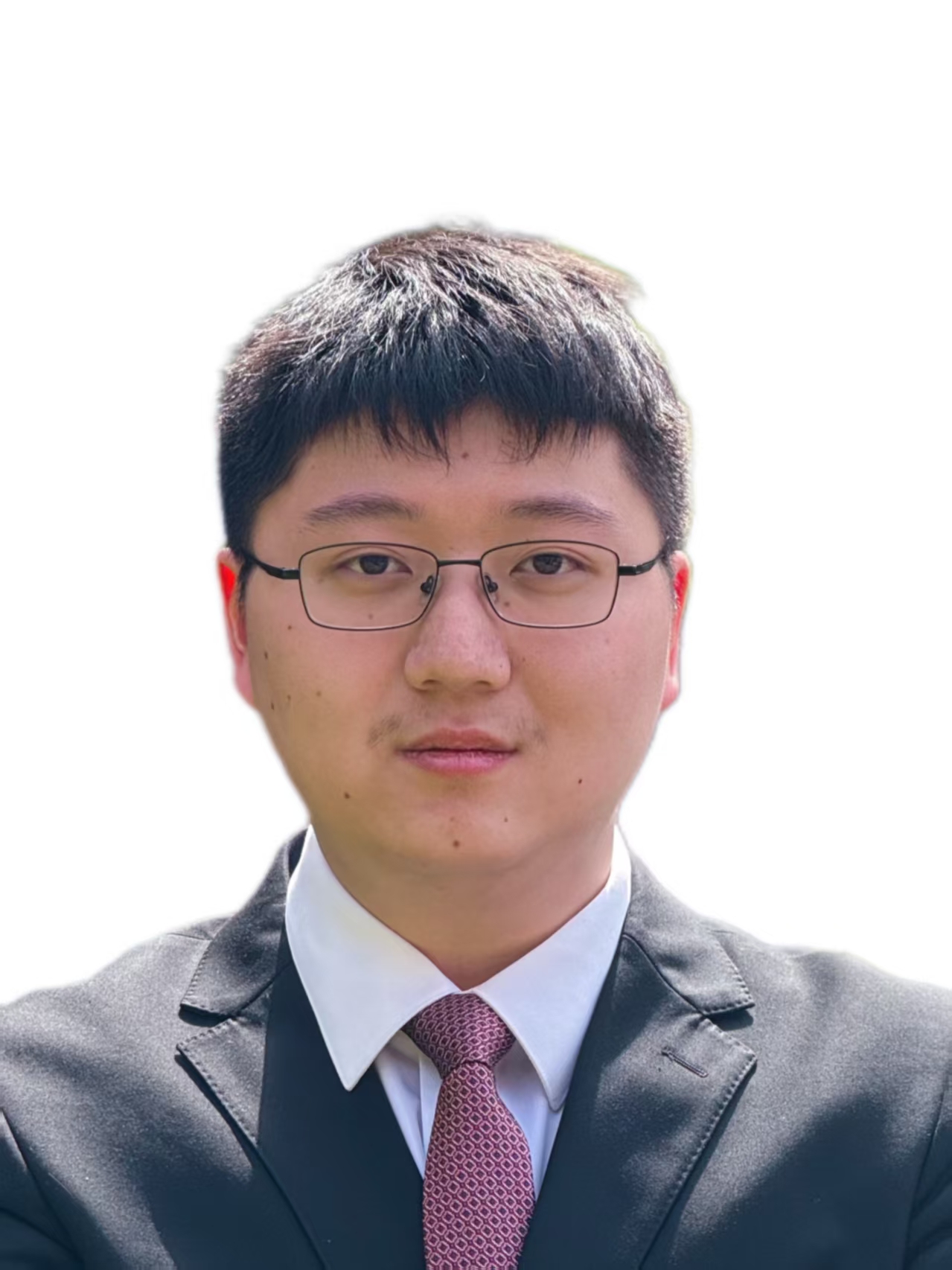}}]{Huayu Zheng} received the B.S. degree in Artificial Intelligence (Top Talent Pilot Program) from Shanghai Jiao Tong University, Shanghai, China, School of Electronic Information and Electrical Engineering in 2025. He is currently working toward the Ph.D. degree with the School of Computer Science, Shanghai Jiao Tong University. His research interests lie in Image and Video Generation and Control.
\end{IEEEbiography}

\begin{IEEEbiography}[{\includegraphics[width=1in,height=1.25in,clip,keepaspectratio]{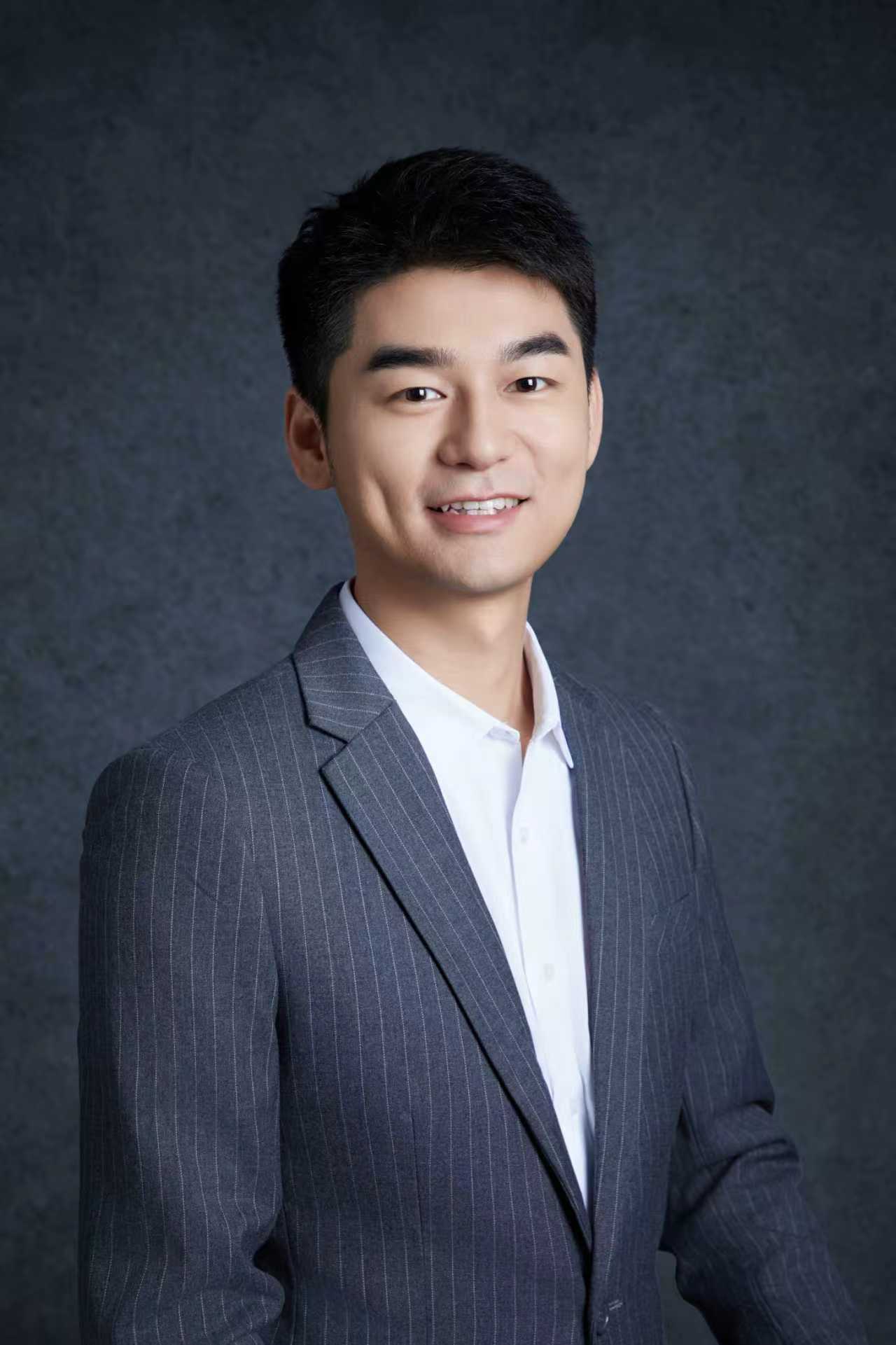}}]{Lei Bai} (Member, IEEE) received the PhD degree from UNSW, Sydney, in 2021. He is currently a Research Scientist at Shanghai Artificial Intelligence Laboratory. His research interests include Science Intelligence, Autonomous Discovery, Multi-Agents System, and their applications in cross-disciplines. Lei has authored or coauthored more than 100 papers in top-tier AI conferences and journals, including Nature Machine Intelligence, Nature Communications, Science Advances, IEEE Transactions on Pattern Analysis and Machine Intelligence, NeurIPS, ICML, and CVPR. He also regularly serves as a program committee member or reviewer for these journals and conference. Lei served as an Area Chair for PRCV 2023/2024 and a Workshop Chair for DICTA 2022 and VALSE 2023. He was the recipient of the 2024 IEEE TCSVT Best Paper Award, 2022 WAIC Yunfan Award, and 2019 Google PhD Fellowship.
\end{IEEEbiography}

\begin{IEEEbiography}[{\includegraphics[width=1in,height=1.25in,clip,keepaspectratio]{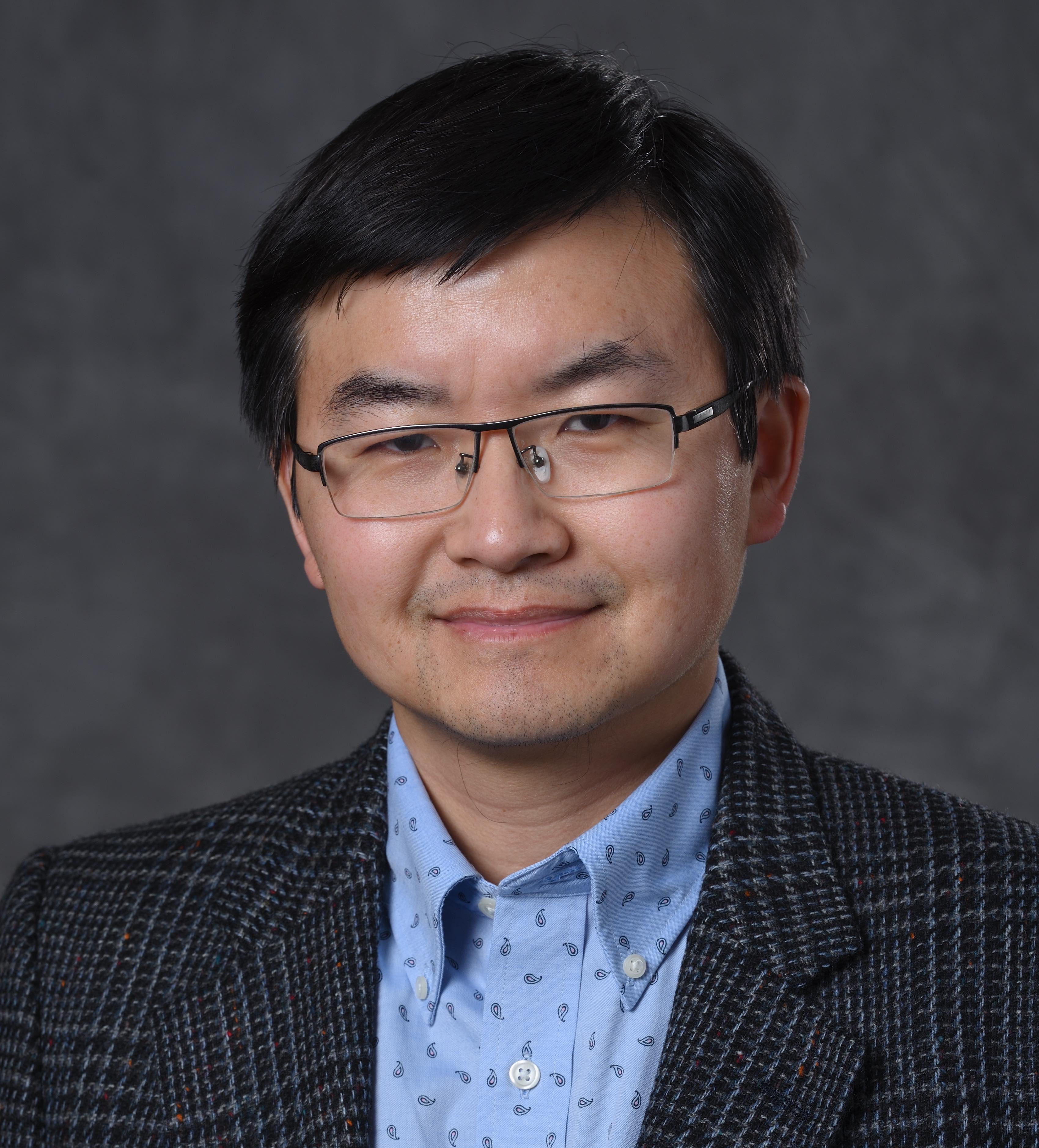}}]{Xiaoming Liu} (Fellow, IEEE) is the MSU Foundation Professor, and Anil and Nandita Jain Endowed Professor at the Department of Computer Science and Engineering of Michigan State University (MSU). He received Ph.D. degree from Carnegie Mellon University in 2004. Before joining MSU in 2012 he was a research scientist at General Electric (GE) Global Research. He works on computer vision, machine learning, and biometrics especially on face related analysis and 3D vision. Since 2012 he helps to develop a strong computer vision area in MSU who is ranked top 15 in US according to csrankings.org. He is an Associate Editor of IEEE Transactions on Pattern Analysis and Machine Intelligence. He has authored more than 200 scientific publications, and has filed 35 U.S. patents. His work has been cited over 30k times according to Google Scholar, with an H-index of 84. He is a fellow of The Institute of Electrical and Electronics Engineers (IEEE) and International Association for Pattern Recognition (IAPR).
\end{IEEEbiography}

\begin{IEEEbiography}[{\includegraphics[width=1in,height=1.25in, clip,keepaspectratio]{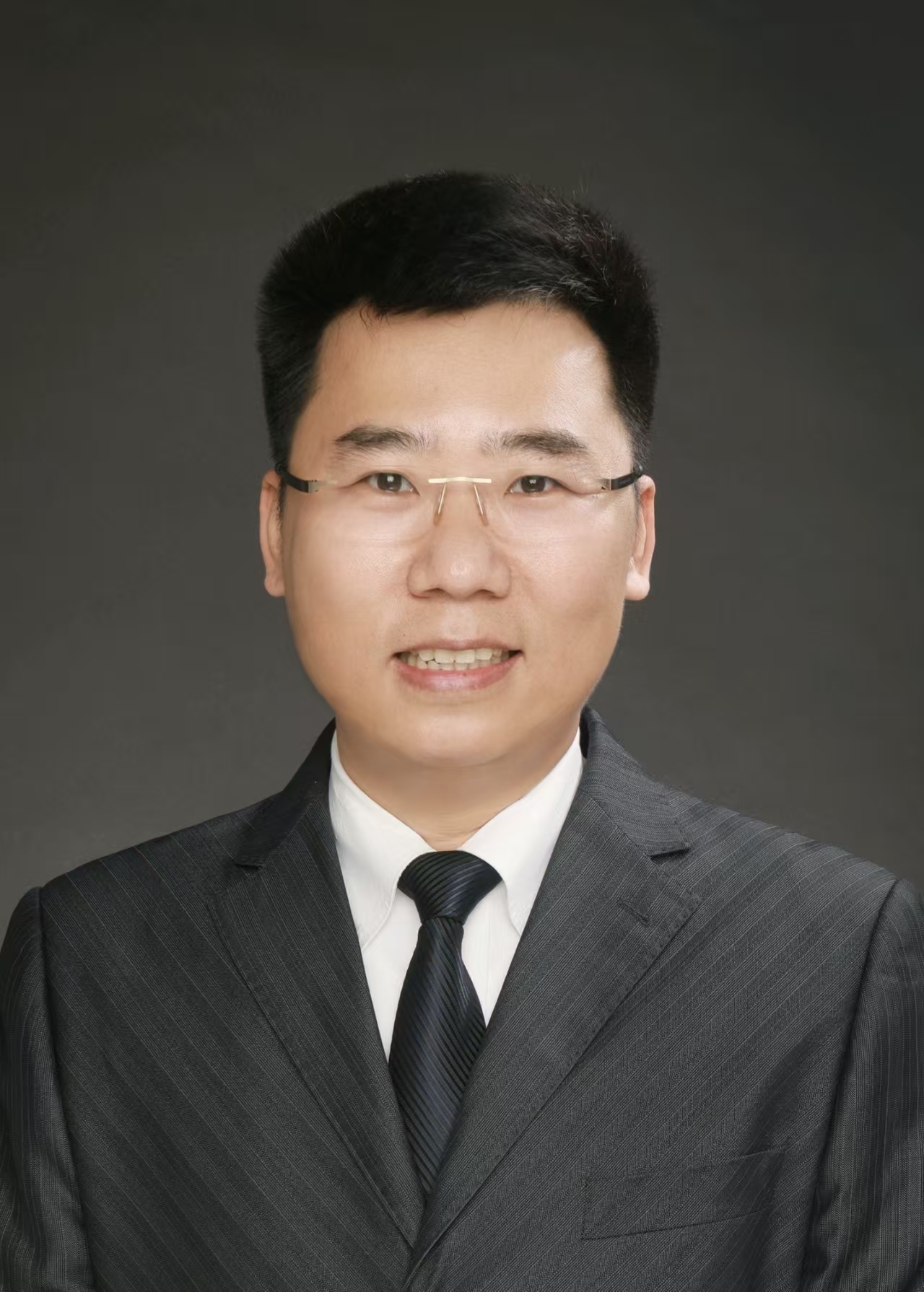}}]{Guangtao Zhai} (Fellow, IEEE) received the B.E. and M.E. degrees from Shandong University, Shandong, China, in 2001 and 2004, respectively, and the Ph.D. degree from Shanghai Jiao Tong University, Shanghai, China, in 2009. From 2008 to 2009, he was a Visiting Student with the Department of Electrical and Computer Engineering, McMaster University, Hamilton, ON, Canada, where he was a Post-Doctoral Fellow, from 2010 to 2012. From 2012 to 2013, he was a Humboldt Research Fellow with the Institute of Multimedia Communication and Signal Processing, Friedrich-Alexander-University of Erlangen–Nüremberg, Germany. He is currently a Professor with the School of Information Science and Electronic Engineering, Shanghai Jiao Tong University. His research interests include multimedia signal processing and perceptual signal processing. He received the Award of National Excellent Ph.D. Thesis from the Ministry of Education of China in 2012.
\end{IEEEbiography}


\end{document}